\documentclass[lettersize,journal]{IEEEtran}
\usepackage{amsmath,amsfonts}
\usepackage{algorithmic}
\usepackage{algorithm}
\usepackage{array}
\usepackage[caption=false,font=normalsize,labelfont=sf,textfont=sf]{subfig}
\usepackage{textcomp}
\usepackage{stfloats}
\usepackage{url}
\usepackage{verbatim}
\usepackage{graphicx}
\usepackage{cite}
\newcommand{\wheels}{\includegraphics[width=4mm]{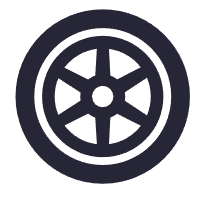}}
\newcommand{\robotdog}{\includegraphics[width=4mm]{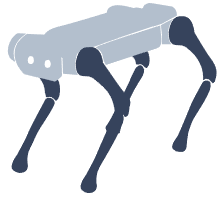}}
\newcommand{\car}{\includegraphics[width=4mm]{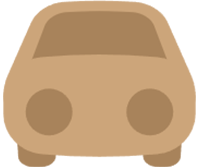}}
\newcommand{\human}{\includegraphics[width=4mm]{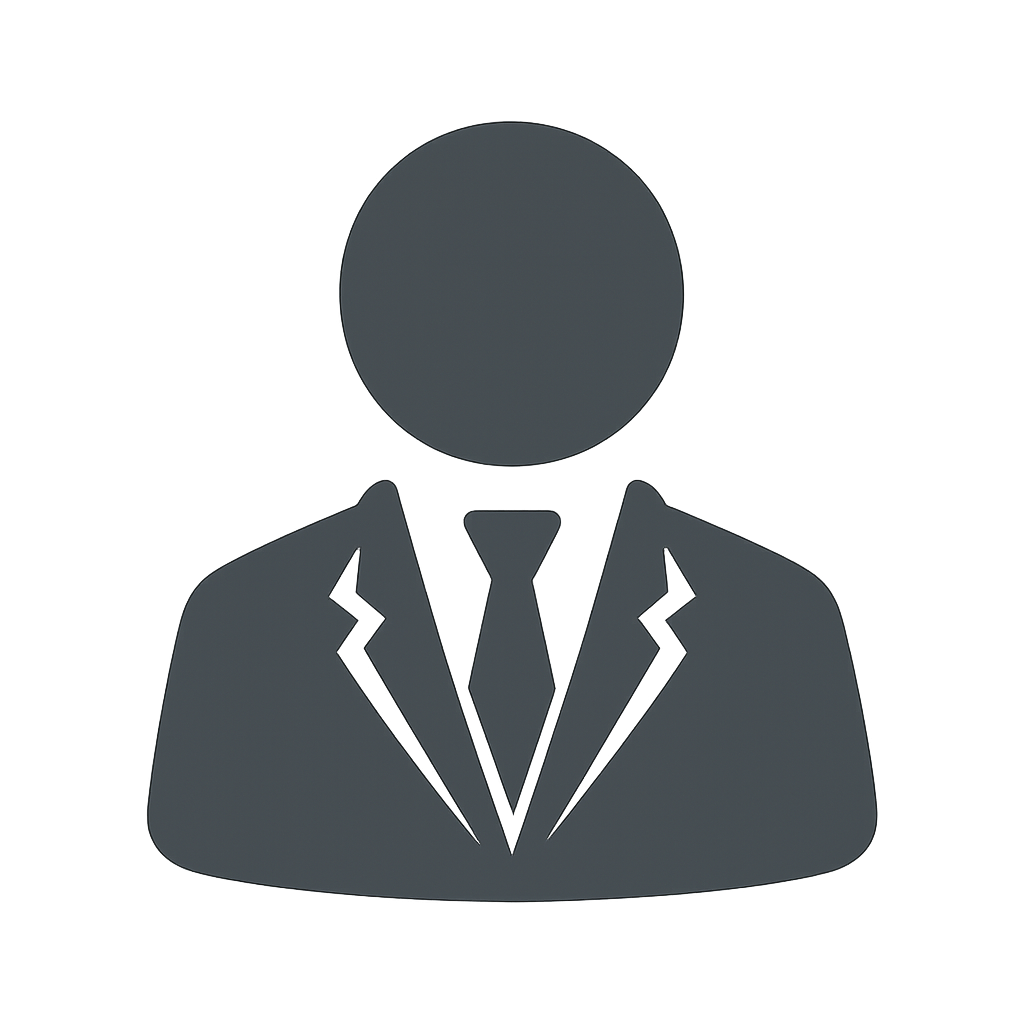}}
\newcommand{\web}{\includegraphics[width=4mm]{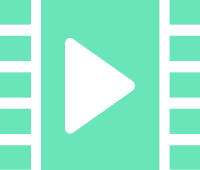}}
\newcommand{\station}{\includegraphics[width=4mm]{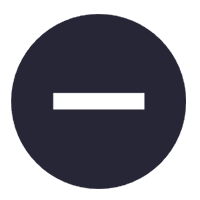}}
\newcommand{\gait}{\includegraphics[width=4mm]{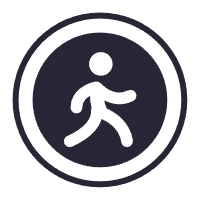}}
\newcommand{\sun}{\includegraphics[width=3.5mm]{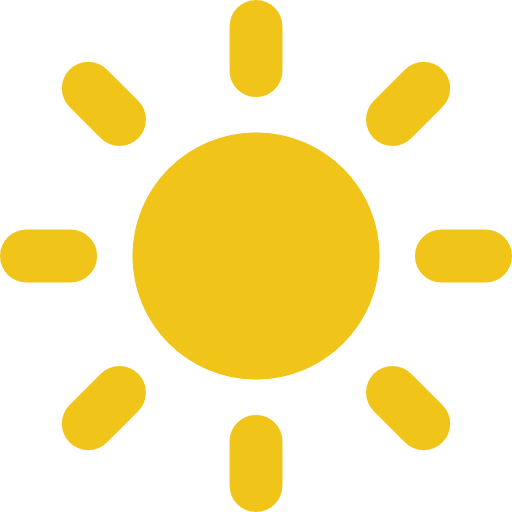}}
\newcommand{\night}{\includegraphics[width=3.5mm]{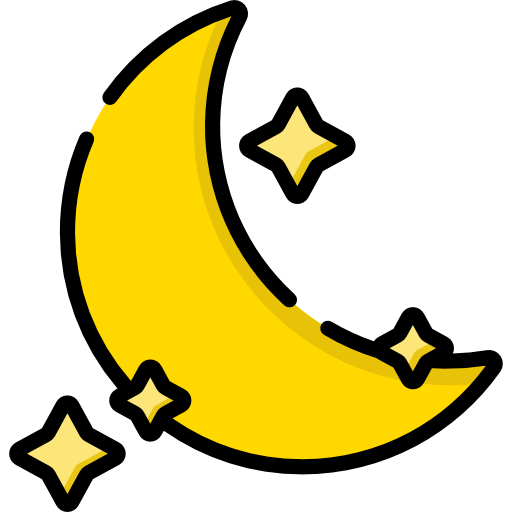}}
\newcommand{\crossmark}{\includegraphics[width=3.5mm]{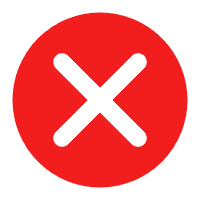}}

\newcommand{\topline}{\noalign{\hrule height 0.8pt}}

\newcommand{\bottomline}{\noalign{\hrule height 0.8pt}}

\usepackage{multirow}
\usepackage{bbding}
\usepackage{adjustbox}
\usepackage{graphicx}
\usepackage{booktabs}
\usepackage{colortbl}
\usepackage{amssymb}
\usepackage[table]{xcolor}

\usepackage[dvipsnames]{xcolor}
\definecolor{hanpurple}{rgb}{0.32, 0.09, 0.98}

\definecolor{table_gray}{RGB}{230,230,230}
\definecolor{table_green}{RGB}{236, 250, 229}

\usepackage[pagebackref=false,breaklinks=true,colorlinks]{hyperref}
\hypersetup{colorlinks,linkcolor={red},citecolor={hanpurple},urlcolor={magenta}}

\begin{document}

\title{Panoramic Multimodal Semantic Occupancy Prediction\\for Quadruped Robots}

\author{Guoqiang Zhao$^{1,*}$, Zhe Yang$^{1,*}$, Sheng Wu$^{1,*}$, Fei Teng$^{1}$, Mengfei Duan$^{1}$, Yuanfan Zheng$^{1}$, Kai Luo$^{1}$,\\and Kailun Yang$^{1,2,\dag}$% <-this % stops a space
\thanks{This work was supported in part by the National Natural Science Foundation of China (Grant No. 62473139), in part by the Hunan Provincial Research and Development Project (Grant No. 2025QK3019), and in part by the State Key Laboratory of Autonomous Intelligent Unmanned Systems (the opening project number ZZKF2025-2-10).}
\thanks{$^{1}$The authors are with the School of Artificial Intelligence and Robotics, Hunan University, Changsha 410012, China (e-mail: kailun.yang@hnu.edu.cn).}
\thanks{$^{2}$The author is also with the National Engineering Research Center of Robot Visual Perception and Control Technology, Hunan University, Changsha 410082, China.}
\thanks{$^{*}$Equal contribution.}
\thanks{$^{\dag}$Corresponding author: Kailun Yang.}
}

\maketitle

\begin{abstract}
Panoramic imagery provides holistic 360{\textdegree} visual coverage for environmental perception in quadruped robots. 
However, existing occupancy prediction methods are primarily designed for wheeled autonomous driving and rely heavily on RGB cues, which limits their robustness in complex, dynamically changing environments. 
To bridge this gap, we introduce PanoMMOcc, the first real-world panoramic multimodal occupancy dataset for quadruped robots, comprising four sensing modalities collected across diverse scenes. 
We further propose VoxelHound, a panoramic multimodal occupancy perception framework tailored to legged locomotion and spherical imaging. 
VoxelHound incorporates a \textbf{Vertical Jitter Compensation (VJC)} module to mitigate severe viewpoint perturbations caused by body pitch and roll during locomotion, enabling more consistent spatial reasoning, and a \textbf{Multimodal Information Prompt Fusion (MIPF)} module to effectively integrate panoramic visual cues with auxiliary modalities for enhanced volumetric occupancy prediction. 
We also establish a comprehensive benchmark on PanoMMOcc and provide detailed dataset analyses to enable systematic evaluation in challenging embodied perception scenarios.
Extensive experiments demonstrate that VoxelHound achieves state-of-the-art performance on PanoMMOcc, with a \textbf{+4.16} gain in mIoU.
The dataset and code will be publicly released to facilitate future research on panoramic multimodal 3D perception for embodied robotic systems at \href{https://github.com/SXDR/PanoMMOcc}{PanoMMOcc}, along with the calibration tools released at \href{https://github.com/losehu/CameraLiDAR-Calib}{CameraLiDAR-Calib}.

\end{abstract}


\section{Introduction}
\label{sec:intro}
\IEEEPARstart{P}{anoramic} perception is emerging as a critical capability for embodied intelligence~\cite{gao2022review,lin2025one,ai2025survey}, enabling a holistic and continuous understanding of the surrounding environment. Compared with conventional narrow Field-of-View (FoV) cameras, panoramic imaging provides full 360{\textdegree} observation without blind spots, which is particularly advantageous for mobile agents operating in dynamic and unstructured scenes~\cite{wei2024onebev,wenke2025dur360bev,zhang2025humanoidpano,yang2020ds,yang2019can,kim2024pair360}. 
As embodied platforms increasingly interact with complex real-world scenes, 
panoramic vision becomes essential for perception-driven decision-making~\cite{luo2025omnidirectional,yang2019pass,chen2024360+,dong2024panocontext}.
However, panoramic images remain 2D projections of a 3D world and lack explicit spatial structure. For embodied agents that require reliable navigation and interaction, 3D scene reasoning is essential.
Occupancy representation has recently gained attention as an effective intermediate representation, modeling free, occupied, and unknown space in a unified form~\cite{huang2023tpvformer,yu2023flashocc,yang2025daocc,wu2025embodiedocc,wang2025embodiedocc++,li2025voxdet,oh2025_3d_prototype,wang2023openoccupancy,li2025occmamba,zhu2026visual_geometry_evidence}. 
By bridging perception and motion planning, occupancy prediction enables spatial reasoning under partial observability and supports safe and efficient robot behavior.

\begin{figure}[t!]
    \centering
    \includegraphics[width=\linewidth]{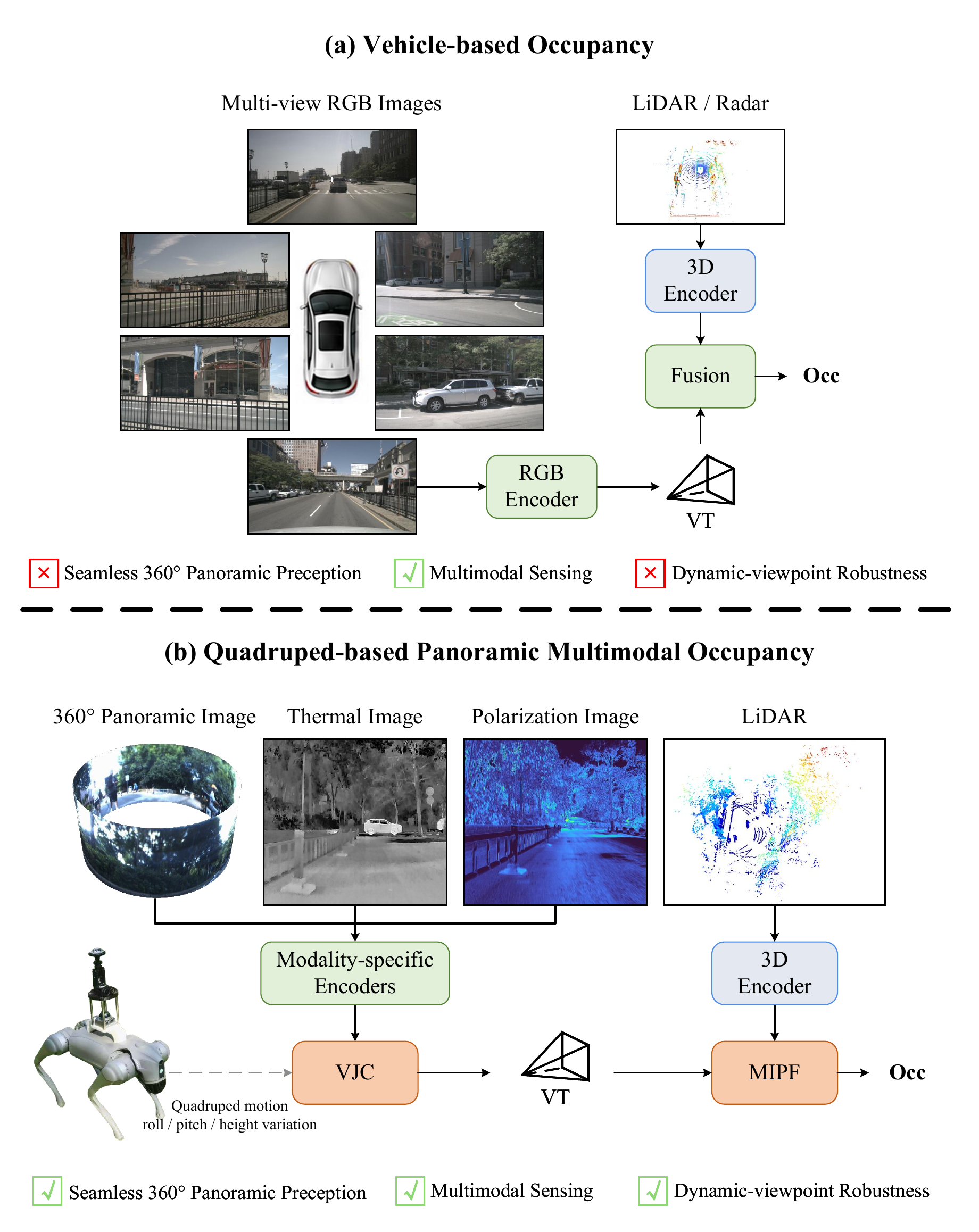}
    \vskip-2ex
    \caption{Comparison of vehicle-based and quadruped-based multimodal occupancy perception. Quadruped-based perception provides seamless 360{\textdegree} panoramic sensing, where VJC mitigates motion-induced viewpoint changes and MIPF performs efficient multimodal information fusion.}

    \vskip-2ex
    \label{fig:motivation}
\end{figure}

Deploying panoramic occupancy perception on real robotic platforms introduces significant challenges, particularly for quadruped robots. Compared with wheeled platforms (Fig.~\ref{fig:motivation}a), quadruped platforms (Fig.~\ref{fig:motivation}b) have low sensor viewpoints, frequent self-occlusions, and strong ego-motion induced by gait dynamics~\cite{wu2025quadreamer,luo2025omnidirectional}. 
Although panoramic cameras mitigate view limitations, relying solely on the RGB modality remains fragile under illumination variations, low-texture regions, and long-range perception scenarios. 
Robust panoramic occupancy perception thus calls for multimodal sensory integration. 
Different sensing modalities provide complementary cues for panoramic scene understanding~\cite{ma2024licrocc,pan2024co_occ,ming2024occfusion}.
RGB images offer rich semantic information~\cite{pan2024co_occ}, thermal sensing enhances robustness under low-light conditions~\cite{ha2017mfnet}, polarization imaging reveals material-dependent and weak-target cues~\cite{xiang2021polarization}, and LiDAR supplies accurate geometric and depth measurements~\cite{ma2024licrocc}.
Integrating these complementary modalities into a unified framework enables more reliable and comprehensive 3D scene modeling for embodied intelligence systems.

Despite this need, existing datasets remain inadequate for this research direction. 
Current panoramic datasets mainly target 2D tasks, such as detection and segmentation, and lack 3D occupancy annotations. 
In contrast, existing occupancy benchmarks are largely designed for autonomous driving, relying on multi-view cameras and vehicle-mounted LiDAR, and rarely consider panoramic imaging or quadruped platforms. 

To bridge this gap, we present a panoramic multimodal occupancy benchmark collected on a real quadruped robot. The dataset provides synchronized panoramic, thermal, polarization, and LiDAR data, along with dense occupancy annotations across diverse environments and realistic robot motions. 
Furthermore, we introduce a baseline panoramic multimodal occupancy prediction framework to establish reference performance and highlight the unique challenges posed by this benchmark. 
Together, our benchmark and baseline provide a systematic foundation for future research on panoramic multimodal occupancy perception in embodied intelligence.
Extensive experiments on PanoMMOcc show that VoxelHound achieves state-of-the-art performance at 23.34 mIoU, surpassing the best camera-only MonoScene (8.94; +14.40) and the best multimodal EFFOcc-T (19.18; +4.16, +21.7\%).

At a glance, we deliver the following contributions: 
\begin{itemize}
    \item We introduce PanoMMOcc, the first panoramic multimodal occupancy dataset for quadruped robots, featuring 360{\textdegree} panoramic, thermal, polarization, and LiDAR data, along with dense occupancy annotations collected under realistic robot motions and diverse scenes.
    \item We propose VoxelHound, the first panoramic multimodal semantic occupancy framework, featuring a Vertical Jitter Compensation (VJC) module to mitigate ego-motion distortions and a Multimodal Information Prompt Fusion (MIPF) module for efficient multimodal feature fusion.
    \item Extensive benchmarking demonstrates that VoxelHound establishes state-of-the-art performance on PanoMMOcc, validating its effectiveness and robustness for panoramic multimodal occupancy perception on quadruped robots.
\end{itemize}
    
\section{Related Work}
\label{sec:related_work}

\subsection{Panoramic Scene Understanding}
Panoramic images provide full 360{\textdegree} environmental coverage and have attracted growing attention in autonomous driving and robotics. 
Existing research can be broadly grouped into semantic perception and geometric reasoning tasks~\cite{lin2025one,ai2025survey,zhu2026panoramic_scene_analysis,meng2025_3d_scene_geometry}.
For semantic understanding, early methods unfold panoramic images with equirectangular projection (ERP) and divide them into several segments for independent encoding and subsequent fusion~\cite{yang2019pass,yang2020ds}. 
To mitigate the domain gap between perspective and panoramic imagery, a large body of work explores unsupervised domain adaptation strategies, including explicit adversarial learning~\cite{ma2021densepass,zheng2023both,zheng2024360sfuda++}, implicit prototypical adaptation~\cite{zhang2022bending,zhang2024behind,zheng2024semantics,zheng2023look_neighbor}, and pseudo-label generation as supervision signals~\cite{zhang2024goodsam,zhong2025omnisam,zhang2024goodsam++}. 
Recent work further improves source-free adaptation through pseudo-label denoising and cross-resolution alignment~\cite{chang2026denoise}.
In addition, several specialized designs are introduced, \textit{e.g.}, distortion convolutions~\cite{hu2022distortion,orhan2022semantic}, spherical deformable patch embedding~\cite{li2023sgat4pass,xu2025mamba4pass}, transformer-based perception modules~\cite{cao2024geometric,lan2025deformable,guttikonda2024single}, and spherical projection transformation~\cite{tan2025dasc} to learn the distorted features in panoramic data. 
Beyond pixel-wise semantics, panoramic perception also extends to Bird's-Eye-View (BEV) mapping and geometric structure estimation. 
BEV-oriented approaches transform fisheye or panoramic inputs into top-down representations for scene understanding~\cite{samani2023f2bev,yogamani2024fisheyebevseg,li2025fishbev,liu2025articubevseg,wenke2025dur360bev,wei2024onebev,teng2024360bev}, with recent work leveraging LiDAR to camera knowledge distillation for single panoramic camera BEV segmentation~\cite{sun2026kd360}.
Panoramic layout estimation methods instead focus on recovering indoor 3D structures through projection-aware reasoning and geometric disentanglement~\cite{fernandez2018layouts,yang2019dula,shen2023disentangling,shen2024360,su2023gpr}.

Despite these advances, existing work primarily targets 2D semantic or specific geometric tasks. Occupancy modeling from panoramic inputs remains largely unexplored, especially for real quadruped robots with multimodal sensing.

\subsection{3D Semantic Occupancy Prediction}

Semantic occupancy prediction aims to assign semantic labels to 3D voxels, providing a dense and fine-grained representation of the surrounding environment.

\textbf{Camera-centric methods.}
Due to the low cost and rich semantics of RGB images, camera-based occupancy prediction receives significant attention. Early work, such as MonoScene~\cite{cao2022monoscene}, explores semantic scene completion from a single image, while subsequent approaches extend to multi-camera surround-view settings with transformer-based and view transformation mechanisms~\cite{huang2023tpvformer,wei2023surroundocc,li2023voxformer,yu2023flashocc,huang2024gaussianformer}. 
To improve efficiency or reduce annotation dependence, recent studies investigate self-supervised learning, compact voxel querying, coarse-to-fine reasoning, and instance-level modeling~\cite{huang2024selfocc,ma2024cotr,li2025voxdet,oh2025_3d_prototype}. 
Beyond autonomous driving, emerging studies have begun to investigate occupancy perception for embodied robotic platforms.
OneOcc~\cite{shi2025oneocc} introduces a vision-only panoramic occupancy framework for legged robots, while other methods study stereo occupancy for humanoids and monocular occupancy in sidewalk scenes~\cite{guo2026humanoid_omniocc,ma2026monocular}.

\textbf{Multimodal methods.}
To overcome the limitations of single-modality perception, multimodal methods combine cameras with LiDAR or radar for more reliable 3D scene understanding~\cite{wang2023openoccupancy,ming2025occfusion,pan2024co_occ,wang2024occgen,yang2025daocc,li2025occmamba,duan2025sdgocc,lv2026gau_occ,zheng2026doracamom}. 
Instead of direct fusion, several methods transfer geometric knowledge across modalities via cross-modal distillation~\cite{ma2024licrocc,wang2025l2cocc,li2026occdistill}.

Overall, existing occupancy methods are predominantly designed for autonomous driving with multi-view perspective cameras, while quadruped-oriented approaches remain camera-only, leaving panoramic multimodal occupancy perception under dynamic legged locomotion largely unexplored.

\begin{figure*}[t!]
    \centering
    \includegraphics[width=0.98\linewidth]
    {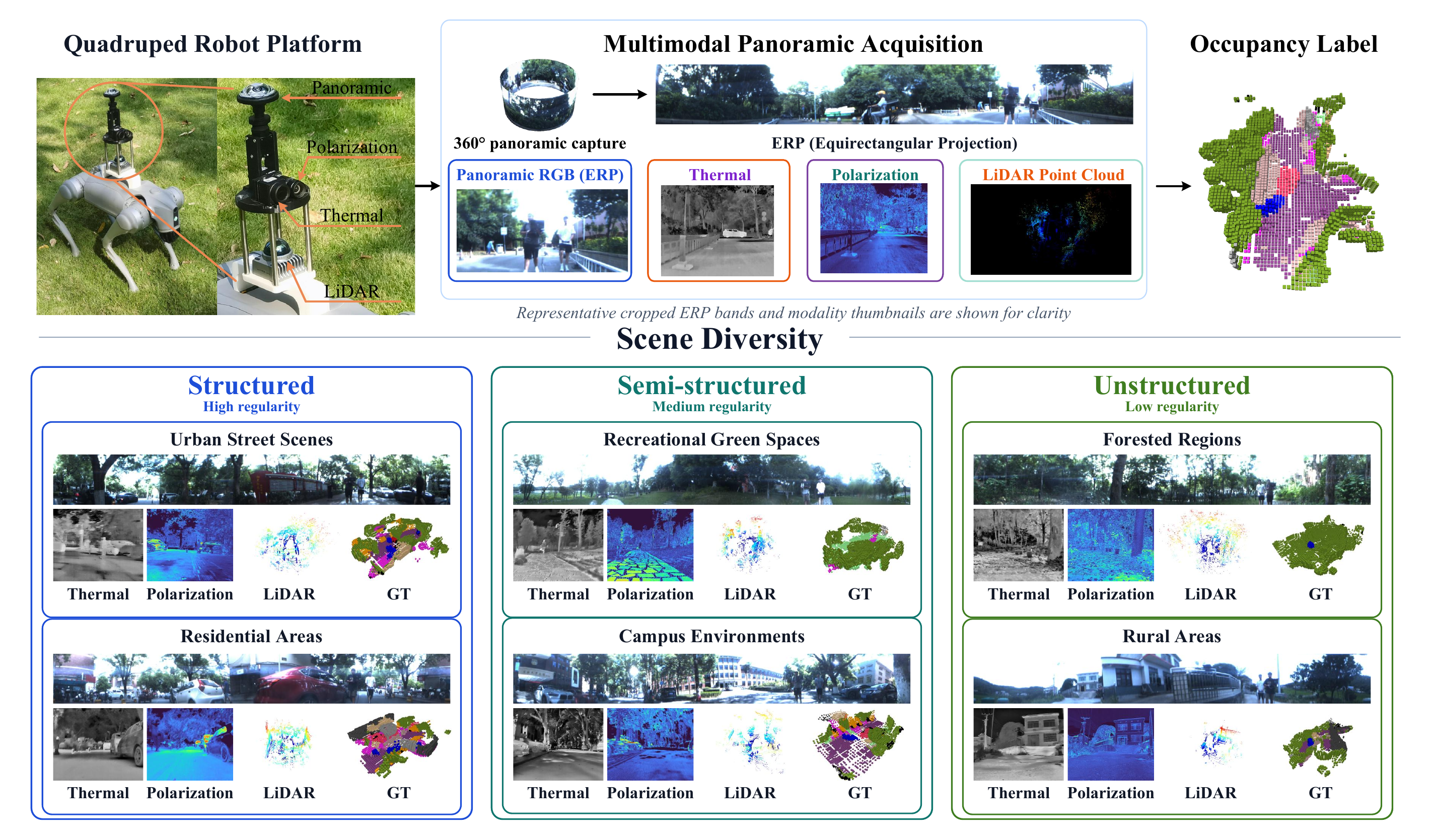}
    \vskip-2ex
    \caption{Overview of the proposed quadruped-based panoramic multimodal occupancy dataset, integrating  panoramic RGB, thermal, polarization, and LiDAR sensing with corresponding 3D occupancy annotations across six scene types, ranging from structured to unstructured environments.}
    
    \label{fig:datashow}
    \vskip-2ex
\end{figure*}

\subsection{Multimodal Occupancy Prediction Datasets}
Comprehensive scene perception relies on high-quality data. 
In recent years, with the continuous advancement of sensor performance and data acquisition technologies, an increasing number of multimodal datasets have been introduced.
Tab.~\ref{tab:datasets} summarizes the publicly available multimodal semantic occupancy prediction datasets.
Early benchmarks such as SemanticKITTI~\cite{behley2019semantickitti} provide LiDAR-based annotations with limited camera coverage. 
Subsequent works extend large-scale driving datasets to support occupancy learning, including SurroundOcc~\cite{wei2023surroundocc}, Occ3D-nuScenes~\cite{tian2023occ3d}, and OpenOccupancy~\cite{wang2023openoccupancy}, which offer annotations for camera-centric and camera–LiDAR multimodal methods.
Beyond conventional camera-LiDAR setups, RadarOcc~\cite{ding2024radarocc} incorporates radar sensing, whereas WildOcc~\cite{zhai2024wildocc} targets off-road environments with monocular camera–LiDAR pairs.
OmniHD-Scenes~\cite{zheng2024omnihd} further integrates multi-camera, LiDAR, and radar sensors for omnidirectional perception in driving scenarios.
QuadOcc~\cite{shi2025oneocc} provides real-world panoramic occupancy data collected by a quadruped robot, but supports only RGB-based perception.

Overall, existing datasets mainly focus on autonomous driving with perspective cameras, whereas quadruped occupancy datasets remain limited to a single visual modality. 
To move beyond this setting, we establish PanoMMOcc to enable holistic and reliable occupancy perception for quadruped robots through panoramic RGB and complementary modalities.

\begin{table*}[t!]
	\caption{Typical datasets for Semantic Occupancy Prediction. Abbreviations: \car\ (Car),\human\ (Human), \robotdog\ (Quadruped Robot), \web\ (Internet images/videos), \wheels\ (Wheels), \gait\ (Gait), \station\ (Stationary), \sun\ (Day), \night\ (Night), \crossmark\ (Inapplicable), Plat. (Platform), Move. (Movement), n.a. (Not applicable),  C (Camera), D (Depth), E (Event), L (LiDAR), R (Radar), T (Thermal), and P (Polarization).}
	\label{tab:datasets}
    \vskip-2ex
	\centering
    %\scriptsize
	%\tabcolsep=10pt
         \setlength{\tabcolsep}{5pt}
	\renewcommand\arraystretch{1}
	\begin{tabular}{c|cccc|ccc|ccc}
		%\toprule%第一道横线
        \topline
        %\rowcolor{table_gray}
		&\multicolumn{4}{c|}{Data}&\multicolumn{3}{c|}{Domain}&\multicolumn{3}{c}{Annotation} \\
        %\rowcolor{table_gray}
		\multirow{-2}{*}{Dataset}&Year&Scene&Camera&Modality&Plat.&Move.&Lighting&Classes&Voxel Grid&Frames \\

		%\midrule%第二道横线 
        \hline
        \hline

        \rowcolor{table_gray}
        \multicolumn{11}{c}{\textit{Indoor scene completion}}\\
        \hline

            MonoScene-NYUv2~\cite{cao2022monoscene} &2022&Indoor&Pinhole&C+D
        &\web&\station&\crossmark&11 &$240{\times}240{\times}144$&1.4K\\

        Occ-ScanNet~\cite{yu2024monocular_occupancy}&2024&Indoor&Pinhole&C+D
        &\web&\station&\crossmark
        &11&$60{\times}60{\times}36$&65.5K\\

            EmbodiedOcc-ScanNet~\cite{wu2025embodiedocc}&2025&Indoor&Pinhole&C+D
        &\web&\station&\crossmark
        &11&-&20.2K \\

            \rowcolor{table_gray}
        \multicolumn{11}{c}{\textit{Autonomous driving and wheeled platforms}}\\
        \hline

            SemanticKITTI~\cite{behley2019semantickitti}&2019&Outdoor&Pinhole&C+L&\car&\wheels&\sun&19& 256${\times}$256${\times}$32&9K\\
            SemanticPOSS~\cite{pan2020semanticposs}&2020&Outdoor&n.a.  &L&\car&\wheels&\sun&14&256${\times}$256${\times}$32&3K\\
            SurroundOcc~\cite{wei2023surroundocc}&2023&Outdoor&Pinhole&C&\car&\wheels&\sun\night&16&200${\times}$200${\times}$16&34K\\
            Occ3D-nuScenes~\cite{tian2023occ3d}&2023&Outdoor&Pinhole&C&\car&\wheels&\sun\night&16&200${\times}$200${\times}$16&40K\\
            OpenOcc~\cite{tong2023scene_as_occupancy}&2023&Outdoor&Pinhole&C&\car&\wheels&\sun\night&16&200${\times}$200${\times}$16&40K\\
            OpenOccupancy~\cite{wang2023openoccupancy}&2023&Outdoor&Pinhole&C+L&\car&\wheels&\sun\night&16&512${\times}$512${\times}$40&34K\\
            RadarOcc~\cite{ding2024radarocc}&2024&Outdoor&n.a.&R&\car&\wheels&\sun\night&2&128${\times}$128${\times}$14&15K\\

            SSCBench-KITTI-360~\cite{li2024sscbench}&2024&Outdoor&Fisheye&C+L&\car&\wheels&\sun&18&256${\times}$256${\times}$32&13K\\
            WildOcc~\cite{zhai2024wildocc}&2024&Outdoor&Pinhole&C+L&\car&\wheels&\sun&7&100${\times}$100${\times}$40&10K\\
            OmniHD-Scenes~\cite{zheng2024omnihd}&2024&Outdoor&Pinhole&C+L+R&\car&\wheels&\sun\night&11&-&60K\\
            ORAD-3D~\cite{min2025advancing_orad}&2025&Outdoor&Pinhole&C+L&\car&\wheels&\sun\night&10&-&58K\\
            Co3SOP~\cite{wu2025synthetic_v2x}&2025&Outdoor&Pinhole&C+L&\car&\wheels&\crossmark&24&1000${\times}$1000${\times}$70&-\\

            Open-pit Mine~\cite{wu2026unsocc}&2026&Outdoor&Pinhole&C+L&\car&\wheels&\sun\night&11&$256{\times}256{\times}32$&7.4K\\

            DSEC-SSC~\cite{guo2026event} & 2026&Outdoor&Pinhole&C+E&\car&\wheels&\sun\night&14&128${\times}$128${\times}$16&3K \\

            Nuplan-Occ~\cite{li2026scaling} & 2026 &Outdoor&Pinhole&C+L
        &\car&\wheels&\sun\night&10&$400{\times}400{\times}32$&3.6M\\

       \rowcolor{table_gray}
        \multicolumn{11}{c}{\textit{Legged and humanoid platforms}}\\
        \hline

            Human360Occ~\cite{shi2025oneocc} & 2025&Outdoor&Panoramic&C&\human&\gait&\sun\night&10&128${\times}$128${\times}$16&8K \\

            Humanoid Occupancy~\cite{cui2025humanoid}&2025&Indoor&Pinhole&C+L&\human&\gait&\crossmark
        &12&$200{\times}200{\times}24$&40K \\

        Humanoid-OmniOcc~\cite{guo2026humanoid_omniocc} & 2026&Indoor & Pinhole&C+D&\human&\gait&\crossmark&14&384${\times}$384${\times}$44&155K \\
        QuadOcc~\cite{shi2025oneocc}&2025&Outdoor&Panoramic&C&\robotdog&\gait&\sun\night&6&64${\times}$64${\times}$8&24K\\

            \hline
             \rowcolor{table_green}
		\textbf{PanoMMOcc (Ours)}&2026&Outdoor&\textbf{Panoramic}
&\textbf{C+L+T+P}&\robotdog&\gait&\sun\night&\textbf{12}&$\mathbf{64{\times}64{\times}16}$&21.6K\\
        \bottomline
	\end{tabular}
    \vskip-4ex
\end{table*}    

\section{PanoMMOcc Dataset}
To advance panoramic perception in complex real-world environments, we introduce \textbf{PanoMMOcc}, the first panoramic multimodal semantic occupancy prediction dataset collected by a quadruped robot. 
As illustrated in Fig.~\ref{fig:datashow}, PanoMMOcc covers diverse environments, including campuses, urban streets, residential areas, green spaces, rural regions, and forests.
With 360{\textdegree} panoramic coverage and complementary sensing modalities, PanoMMOcc enables robust scene perception under diverse illumination, material, and structural conditions, bridging robotic perception and comprehensive environmental understanding. In the following, we describe the sensor suite, annotation pipeline, and dataset statistics.

\begin{figure}[t]
    \centering
    \includegraphics[width=0.98\linewidth]
    {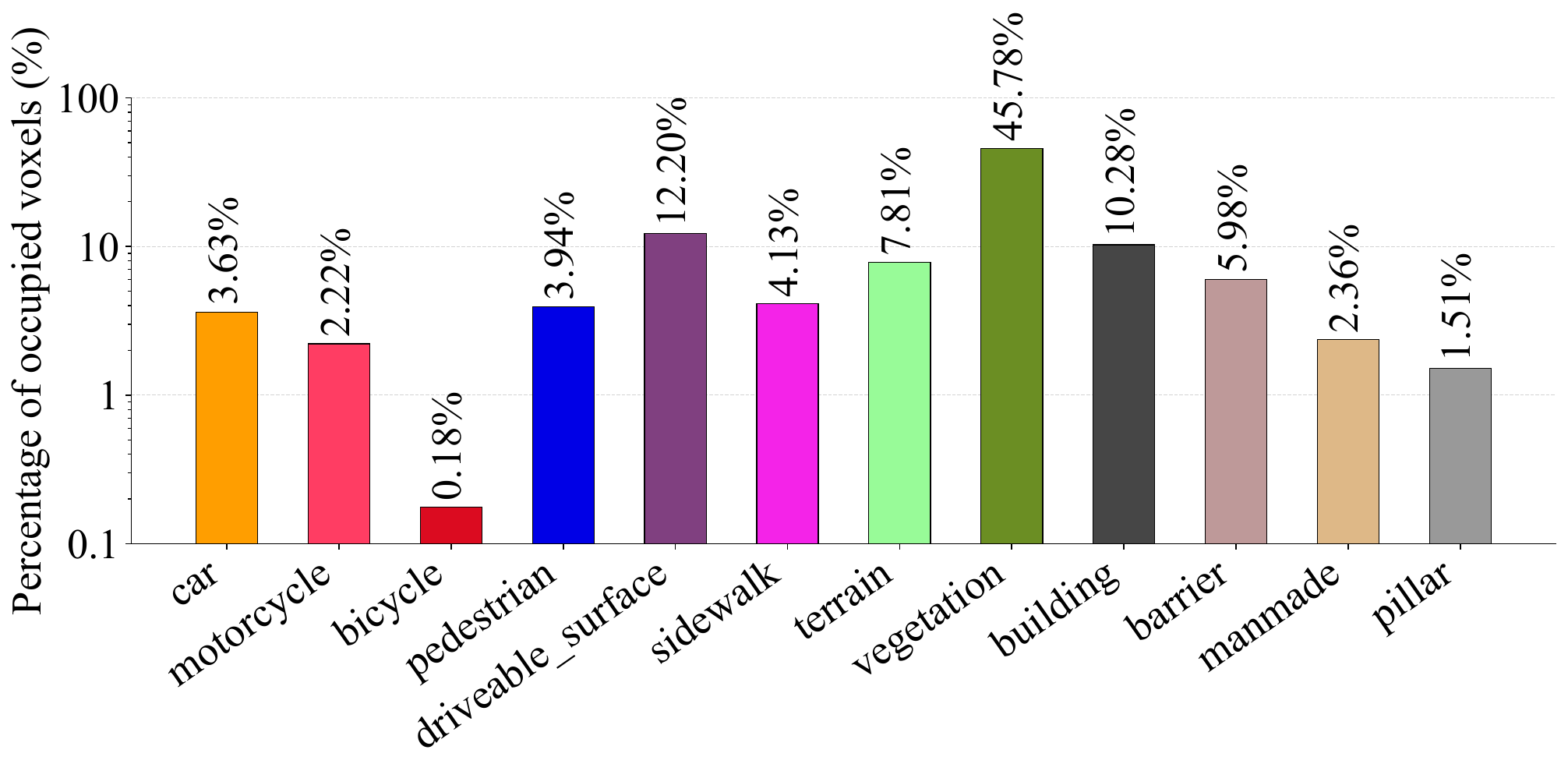}
    \vskip -2ex
    \caption{Semantic class distribution of occupied voxels in the PanoMMOcc.}
    \vskip -4ex
    \label{fig:class_distribution}
\end{figure}

\noindent \textbf{Sensor suite.}
We develop a panoramic multimodal data acquisition platform mounted on a quadruped robot, as shown in Fig.~\ref{fig:datashow}. 
(1) \textit{Quadruped platform.} 
A Unitree Go2 robot is employed for data collection, and its agile locomotion enables traversal of complex outdoor environments.
(2) \textit{Panoramic camera.} 
A Panoramic Annular Lens (PAL) camera provides a $360^{\circ}\times70^{\circ}$ FoV at up to $40$ FPS with a resolution of $2048\times2048$.
(3) \textit{LiDAR.} 
A MID-360 LiDAR mounted on the robot's back captures accurate 3D point clouds.
(4) \textit{Thermal camera.} 
A Guide Infrared N-Driver P302B camera captures thermal images at $25$ FPS with a resolution of $640\times512$.
(5) \textit{Polarization camera.} 
A LUCID TRI050S-QC camera captures polarization images at a resolution of up to $1224\times1024$.

\noindent \textbf{Annotations.}
To obtain high-quality voxel-level labels, we adopt a multi-stage annotation and refinement pipeline.
Professional annotators first provide semantic, object-level, and instance-level annotations to ensure consistency and fine-grained accuracy. 
Each frame is then carefully reviewed to correct semantic inconsistencies, inaccurate object boundaries, and other labeling errors. 
The refined annotations are grouped into $12$ semantic categories to support comprehensive perception in structurally diverse scenes. 
Finally, occupancy labels are generated following the protocol of SurroundOcc~\cite{wei2023surroundocc}, ensuring compatibility with existing benchmarks and enabling standardized comparisons with prior methods.

\noindent \textbf{Statistical analysis.}
PanoMMOcc contains $54$ sequences captured at $10$ Hz, each lasting $40$ seconds, resulting in a total of $21{,}600$ frames. 
Among them, $42$ sequences are selected for semantic annotation to construct the occupancy benchmark. 
Fig.~\ref{fig:class_distribution} presents the class distribution of annotated categories.
The dataset covers both daytime and nighttime conditions, capturing variations in illumination and environmental dynamics.
Each sequence provides synchronized panoramic RGB, thermal, polarization, and LiDAR data, offering comprehensive multimodal observations for panoramic occupancy perception. 
Additional details are provided in the supplementary material.
  
\section{Method}
\subsection{Problem Formulation}
Given a panoramic RGB image $\mathcal{I}^{pal}$, a thermal image $\mathcal{I}^{th}$, a polarization image $\mathcal{I}^{pol}$, and a LiDAR point cloud $\mathcal{P}$, our goal is to predict a dense 3D semantic occupancy grid:
\begin{equation}
\mathbf{O}=\Phi\left(\mathcal{I}^{pal},\mathcal{I}^{th},
\mathcal{I}^{pol},\mathcal{P}\right),
\label{task_formulated}
\end{equation}
where $\Phi$ denotes the multimodal occupancy network and $\mathbf{O}\in\mathbb{R}^{X\times Y\times Z}$ is the predicted voxel grid.

\begin{figure*}[t!]
    \centering
    \includegraphics[width=\linewidth]{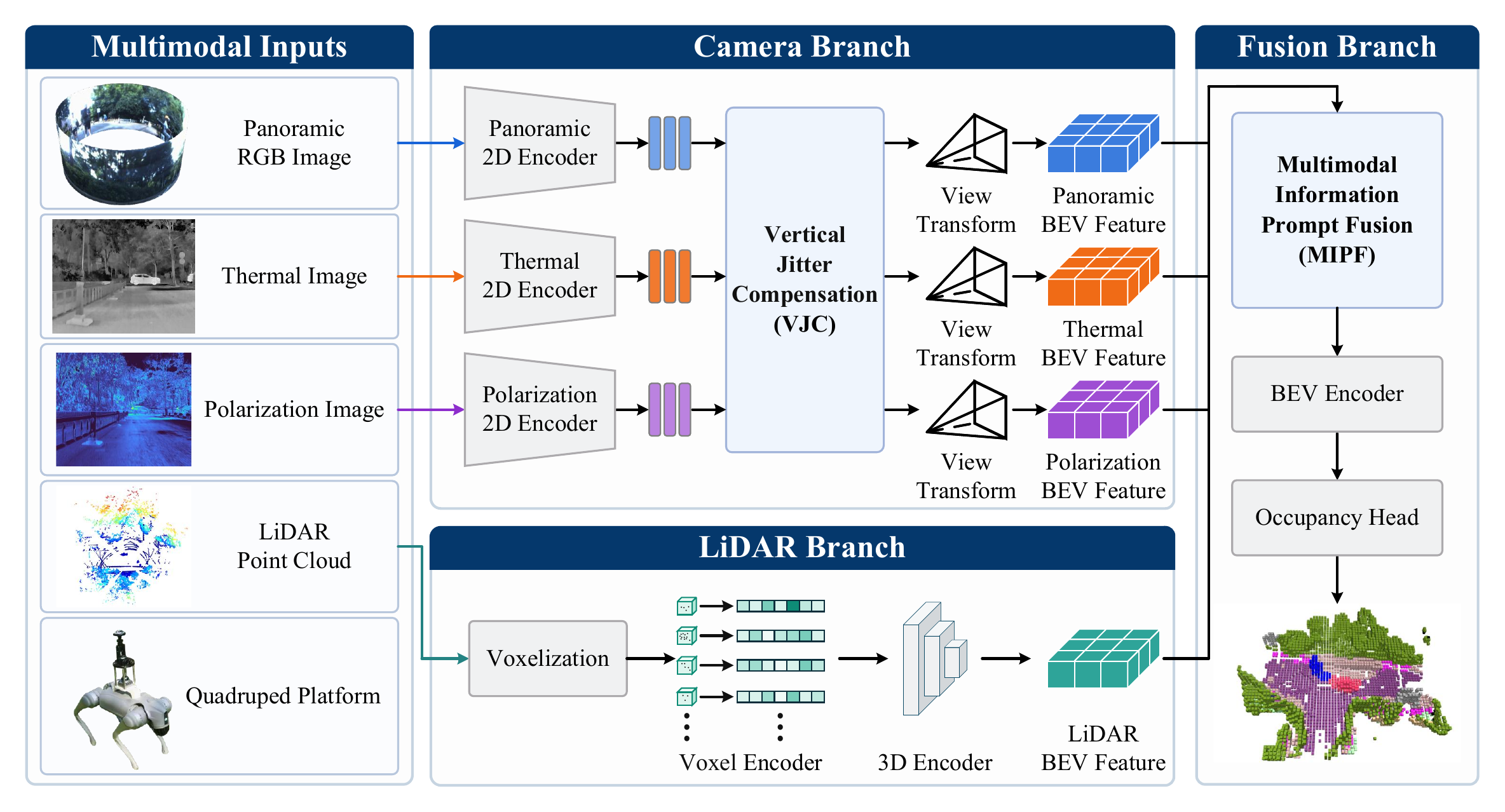} 
    \vskip-2ex
    \caption{Overview of the proposed VoxelHound. The camera branch receives panoramic RGB, thermal, and polarization images, while the LiDAR branch processes raw point clouds. Features from independent encoders are projected into BEV space for multimodal fusion.}
    \label{fig:network}
    \vskip-3ex
\end{figure*}

\subsection{Overview}
Effective 3D scene understanding for quadruped robots requires fully exploiting the complementary cues provided by multimodal sensors. 
Existing occupancy prediction methods often rely on a dominant modality or simple fusion strategies, limiting their robustness under challenging conditions such as illumination changes, material variations, and platform motion.

The overall architecture of VoxelHound is shown in Fig.~\ref{fig:network}. Given a panoramic RGB image, a thermal image, a polarization image, and a LiDAR point cloud, modality-specific encoders first extract their features. Image features are lifted from the 2D image space to BEV space, while LiDAR points are voxelized and projected into the same space, forming unified spatial representations for cross-modal interaction. The aligned BEV features are then fused and refined by a BEV encoder for contextual modeling. Finally, an occupancy head restores the height dimension and produces dense 3D semantic occupancy predictions.
These components form the basic architecture of VoxelHound, as detailed in Sec.~\ref{sec:mfn}.
Furthermore, we introduce two specialized modules: a Vertical Jitter Compensation module (VJC, Sec.~\ref{sec:vjc}) to mitigate locomotion-induced distortions, and a Multimodal Information Prompt Fusion module (MIPF, Sec.~\ref{sec:mipf}) to enhance cross-modal interaction while preserving geometric consistency.

\subsection{Multimodal Fusion Network}
\label{sec:mfn}
Existing multimodal occupancy methods mainly focus on multi-view pinhole RGB cameras and LiDAR or radar sensors~\cite{wang2023openoccupancy,shi2024effocc,ma2024licrocc,ming2025occfusion,yang2025daocc,pan2024co_occ,wang2024occgen,li2025occmamba,duan2025sdgocc}. In contrast, VoxelHound jointly exploits panoramic RGB, thermal, polarization, and LiDAR measurements within a unified BEV representation.

\noindent \textbf{Camera Branch.}
The camera branch processes three image modalities, including panoramic RGB $\mathcal{I}^{pal}$, thermal $\mathcal{I}^{th}$, and polarization $\mathcal{I}^{pol}$. Each modality is independently encoded by a 2D backbone to extract multi-scale features:
\begin{equation}
    {\bf{F}}_{c\_backbone}^m = \left\{ {{\bf{F}}_{1/8}^m,{\bf{F}}_{1/16}^m,{\bf{F}}_{1/32}^m} \right\}, m \in\{pal, th, pol\},
	\label{2d_faeture}
\end{equation}
where ${\bf{F}}_{1/i}$ denotes features at $i\times$ downsampling. 
A Feature Pyramid Network (FPN) aggregates multi-scale features to produce
${\bf{F}}_{c\_neck}^m \in {\mathbb{R}^{{C_{neck}} \times {\frac{H^\prime}{8} } \times {\frac{W^\prime}{8} }}}$.
Subsequently, we apply a 2D-to-BEV view transform to project 2D features into the BEV space. 
As a result, each image modality produces a BEV feature: ${\bf{F}}_{c}^m \in {^{{C_m} \times {H} \times {W}}}, m \in\{pal, th, pol\}$,
which serves as the multimodal camera BEV features for multimodal fusion.

\noindent \textbf{LiDAR Branch.}
Given the LiDAR point cloud $\mathcal{P}$, we voxelize the 3D space within a predefined range. Each voxel retains up to $10$ points, whose mean feature forms a voxel-level representation.
The voxel features are then fed into a sparse 3D convolutional encoder with an overall stride of $8$ to extract hierarchical geometric features.
Next, the sparse 3D features are splatted to BEV features: $ {\bf{F}}_l \in {^{{C_l} \times {H} \times {W}}}$, which serves as the LiDAR BEV feature for multimodal fusion.

\noindent \textbf{Fusion Branch.}
Given the  multimodal camera BEV features ${\bf{F}}_{c}^m \in {^{{C_m} \times {H} \times {W}}}$, $ m \in\{pal, th, pol\}$ and the LiDAR BEV feature ${\bf{F}}_l \in {^{{C_l} \times {H} \times {W}}}$, we first aggregate them through a  fusion module. The fusion process can be formulated as:
\begin{equation}
    \mathbf{F}_{f}=\mathcal{F}_{\text {fusion }}\left(\mathbf{F}_{l}, \mathbf{F}_{c}^{\text {pal }}, \mathbf{F}_{c}^{t h}, \mathbf{F}_{c}^{\text {pol }}\right),
   \label{fusion_bev_faeture}
\end{equation}
where $\mathcal{F}_{\text {fusion}}$ denotes the fusion operator and ${\bf{F}}_f \in {^{{C_f} \times {H} \times {W}}}$ is the fused BEV feature.
The fused features are further refined by a BEV encoder for contextual modeling. Following~\cite{shi2024effocc}, we adopt a SECOND-FPN architecture to enhance multi-scale spatial representation. Finally, an occupancy head predicts the 3D semantic occupancy by reshaping the BEV feature channels into vertical bins, producing
$\mathbf{O} \in \mathbb{R}^{X \times Y \times Z}$.
Each voxel is assigned a semantic label from $0$ to $12$.

\begin{figure*}[t]
    \centering
    \includegraphics[width=\linewidth]{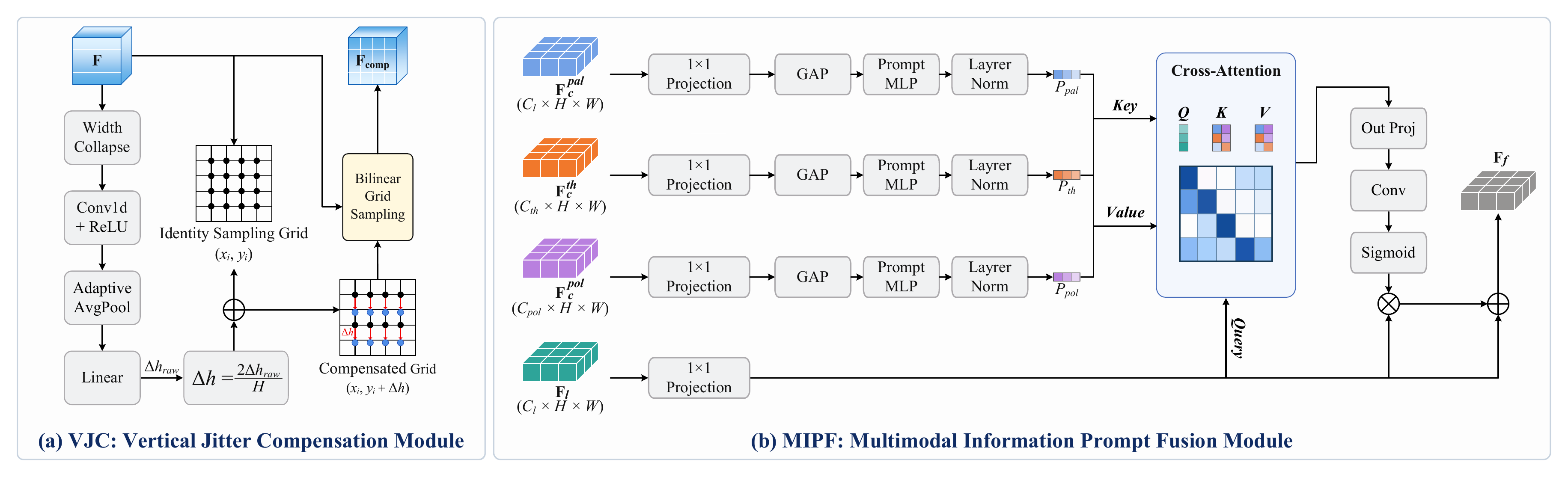} 
    \vskip-2ex
    \caption{Overview of the proposed modules. a) Vertical Jitter Compensation (VJC) module mitigates image distortions caused by quadruped locomotion. b) Multimodal Information Prompt Fusion (MIPF) module for effective multimodal feature fusion.}
    \label{fig:module}
    \vskip-4ex
\end{figure*}

\subsection{Vertical Jitter Compensation}
\label{sec:vjc}
Unlike wheeled autonomous driving platforms, our dataset is collected using a quadruped robot. Due to legged locomotion, body oscillation along the vertical axis introduces vertical jitter in the captured images. 
This gait-induced perturbation causes misalignment in spatial feature representation and degrades the stability of the BEV transformation.
To mitigate this issue, we propose a lightweight Vertical Jitter Compensation (VJC) module, which is inserted between the image encoder and the 2D to BEV view transform stage. 
As illustrated in Fig.~\ref{fig:module}(a), VJC estimates and compensates for vertical feature displacement before lifting image features into the BEV space.

Given an image feature map $\mathbf{F}\in\mathbb{R}^{C\times H\times W}$, VJC first averages the features along the width dimension to obtain a vertical descriptor:
\begin{equation}
    \mathbf{F}_{v}(c, h)=\frac{1}{W} \sum_{w=1}^{W} \mathbf{F}(c, h, w),
\label{vjc_vertical_descriptor}
\end{equation}
where $\mathbf{F}_{v}\in\mathbb{R}^{C\times H}$ summarizes the vertical structure while suppressing horizontal redundancy.
A lightweight Conv encoder $\mathcal{E}{v}$ and a regression head $\mathcal{R}$ then estimate the vertical displacement, which is normalized to the coordinate range used by grid sampling:
\begin{equation}
    \Delta h =
    \frac{2}{H}
    \mathcal{R}\left(\mathcal{E}{v}\left(\mathbf{F}_{v}\right)\right),
\label{vjc_offset}
\end{equation}
where $\mathcal{R}$ consists of adaptive average pooling followed by a linear layer.
Finally, the predicted offset is applied to the vertical coordinate of an identity sampling grid $\mathbf{G}_{0}$, and the compensated feature is obtained by bilinear sampling:
\begin{equation}
    \mathbf{F}_{\text {comp }}=
    \operatorname{GridSample}
    \left(
    \mathbf{F},
    \mathbf{G}_{0}+(0,\Delta h)
    \right).
\label{vjc_compensation}
\end{equation}
The compensated feature $\mathbf{F}_{\text {comp }} \in\mathbb{R}^{C\times H\times W}$ is subsequently fed into the 2D to BEV view transformation.

\definecolor{ncar}{RGB}{255, 158, 0}
\definecolor{nmotorcycle}{RGB}{255, 61, 99}
\definecolor{nbicycle}{RGB}{219, 11, 32}
\definecolor{npedestrain}{RGB}{0, 0, 230}
\definecolor{ntraffic_road}{RGB}{128 ,64 ,128}
\definecolor{nsidewalk}{RGB}{244, 35, 232}
\definecolor{nterrain}{RGB}{152, 251, 152}
\definecolor{nvegetation}{RGB}{107 ,142 ,35}
\definecolor{nbuilding}{RGB}{70, 70, 70}
\definecolor{nbarrier}{RGB}{190 ,153 ,153}
\definecolor{nmanmade}{RGB}{222, 184, 135}
\definecolor{npillar}{RGB}{153, 153, 153}

\captionsetup{font=small}  % 设置标题字体大小为小号

\setlength{\tabcolsep}{1.4pt}  % 列间距
\renewcommand{\arraystretch}{1}  % 行高

\begin{table*}[!t]
\centering
\renewcommand\arraystretch{1}
\caption{Comparison of semantic occupancy prediction methods on the established PanoMMOcc dataset. 
Bold and underlined values indicate the best and second-best results in each column, respectively.
}
\vskip-2ex
\begin{adjustbox}{width=0.95\textwidth,center}
% \resizebox{0.9\textwidth}{!}{
\begin{tabular}{c|c|c|cccccccccccc}
\hline
\noalign{\smallskip}
Methods & Modality & mIoU$\uparrow$ & \rotatebox{90}{\textcolor{ncar}{$\blacksquare$}car}     & \rotatebox{90}{\textcolor{nmotorcycle}{$\blacksquare$}motorcycle}    & \rotatebox{90}{\textcolor{nbicycle}{$\blacksquare$}bicycle} & \rotatebox{90}{\textcolor{npedestrain}{$\blacksquare$}pedestrian} & \rotatebox{90}{\textcolor{ntraffic_road}{$\blacksquare$}driveable surface} & \rotatebox{90}{\textcolor{nsidewalk}{$\blacksquare$}sidewalk} & \rotatebox{90}{\textcolor{nterrain}{$\blacksquare$}terrain} & \rotatebox{90}{\textcolor{nvegetation}{$\blacksquare$}vegetation} & \rotatebox{90}{\textcolor{nbuilding}{$\blacksquare$}building} & \rotatebox{90}{\textcolor{nbarrier}{$\blacksquare$}barrier} & \rotatebox{90}{\textcolor{nmanmade}{$\blacksquare$}mannade} & \rotatebox{90}{\textcolor{npillar}{$\blacksquare$}pillar}    \\
\noalign{\smallskip}
\hline
\noalign{\smallskip}
MonoScene~\cite{cao2022monoscene} &C&8.94&10.56&16.02&0.00&26.81&20.45&2.67&10.06&10.67&4.26&2.92&0.85&1.98\\ 

C-CONet~\cite{wang2023openoccupancy}&C&4.61&0.41&2.69&0.00&21.24&16.86&2.30&5.36&6.08&0.22&0.15&0.00&0.00\\ 
L-CONet~\cite{wang2023openoccupancy} &L&12.48&12.17&10.75&0.00&32.95&23.99&6.17&9.86&23.44&13.79&6.45&0.99&9.17\\ 
M-CONet~\cite{wang2023openoccupancy} &C+L&12.98&11.91&10.96&0.00&30.81&24.54&7.74&11.30&24.43&18.45&6.51&1.53&7.61\\  
OccFusion~\cite{ming2024occfusion}&C&5.47&0.69&8.75&0.00&22.65&14.34&3.62&5.55&7.43&1.07&1.04&0.31&0.25\\
OccFusion~\cite{ming2024occfusion}&L&17.48&15.38&15.71&0.00&38.41&33.00&7.90&16.80&29.69&29.37&12.67&1.09&9.75\\
OccFusion~\cite{ming2024occfusion}&C+L&18.30&15.14&16.56&0.00&38.81&35.53&7.54&19.92&30.09&31.43&12.65&1.85&10.12\\
LiCROcc~\cite{ma2024licrocc}&C&5.02&1.67&5.33&0.00&21.01&16.98&2.39&6.02&4.66&1.42&0.68&0.08&0.00\\
LiCROcc~\cite{ma2024licrocc}&L&17.95&17.05&16.15&0.00&42.52&\underline{36.45}&7.23&16.02&29.18&28.64&10.54&0.18&11.47\\
LiCROcc~\cite{ma2024licrocc}&C+L&18.46&18.65&18.13&0.00&41.91&\textbf{37.25}&4.28&16.86&29.41&30.20&12.04&0.85&11.92\\
EFFOcc-C~\cite{shi2024effocc} &C&4.47&1.68&7.11&0.00&19.90&15.53&0.82&5.30&2.80&0.04&0.47&0.00&0.01\\ 
EFFOcc-L~\cite{shi2024effocc} &L&18.77&20.25&12.50&\textbf{1.28}&44.02&29.47&\underline{8.52}&19.25&28.31&31.79&\underline{16.69}&4.55&8.59\\ 
EFFOcc-T~\cite{shi2024effocc} &C+L&19.18&19.81&12.13&\underline{0.68}&44.16&30.20&7.14&21.49&30.10&33.87&15.18&\underline{5.41}&10.00\\
\hline

&C&5.53&1.34&6.35&0.00&21.03&18.26&2.59&6.91&8.09&0.45&1.03&0.09&0.17\\
&C+T&5.91&3.31&11.09&0.00&19.95&16.80&1.11&8.48&8.28&0.34&1.26&0.30&0.02 \\
&C+P&5.95&1.12&4.19&0.00&20.13&19.59&3.86&9.56&9.68&0.45&2.48&0.14&0.19 \\
&C+T+P&6.14&1.51&8.25&0.03&20.83&18.17&4.47&7.78&8.96&0.08&3.26&0.08&0.21\\ 
&C+L&\underline{22.87}&\textbf{26.86}&\textbf{22.11}&0.00&\underline{48.99}&34.56&6.46&\underline{23.38}&\underline{34.16}&\textbf{37.70}&16.49&\textbf{7.14}&\underline{16.54}\\ 
%\rowcolor{table_green}
\multirow{-6}{*}{VoxelHound (Ours)}&\cellcolor{table_green}C+L+T+P&\cellcolor{table_green}\textbf{23.34}&\cellcolor{table_green}\underline{26.69}&\cellcolor{table_green}\underline{21.77}&\cellcolor{table_green}0.00&\cellcolor{table_green}\textbf{49.53}&\cellcolor{table_green}34.97&\cellcolor{table_green}\textbf{8.59}&\cellcolor{table_green}\textbf{24.61}&\cellcolor{table_green}\textbf{34.41}&\cellcolor{table_green}\underline{37.35}&\cellcolor{table_green}\textbf{18.67}&\cellcolor{table_green}4.71\cellcolor{table_green}&\cellcolor{table_green}\textbf{18.76}\\  %\hline

%\noalign{\smallskip}
\bottomline
\end{tabular}
\label{tab:occ_performance}
\end{adjustbox}
\vskip-4ex
\end{table*}
\setlength{\tabcolsep}{1.4pt}

\subsection{Multimodal Information Prompt Fusion}
\label{sec:mipf}
Direct concatenation or addition treats all sensor modalities equally, although they provide different types of information. 
In practice, heterogeneous sensors contribute fundamentally different information: LiDAR directly measures 3D geometry, while RGB, thermal, and polarization modalities provide complementary semantic and appearance cues. Blindly blending these features may dilute geometric consistency and introduce modality interference.
Thus, MIPF adopts asymmetric fusion, using LiDAR as the geometric basis and image modalities as semantic supplements, as shown in Fig.~\ref{fig:module}(b).

Specifically, MIPF uses the LiDAR BEV feature as spatial queries and compresses each image modality into a compact semantic prompt. We first project all modality features into a shared embedding space:
\begin{equation} 
    \tilde{\mathbf{F}}_{l}=\phi_{l}\left(\mathbf{F}_{l}\right), \quad \tilde{\mathbf{F}}_{c}^m=\phi_{m}\left(\mathbf{F}_{c}^m\right),
\label{mipf_projection}
\end{equation}
where $m\in\{pal, th, pol\}$, and $\phi_l$ and $\phi_m$ denote $1\times1$ convolutional projections.
Each image feature is then globally pooled and processed by a lightweight MLP to generate a modality-specific semantic prompt:
\begin{equation}            
\mathbf{P}_{m}=\mathcal{M}_{m}\left(\operatorname{GAP}\left(\tilde{\mathbf{F}}_{c}^m\right)\right).
   \label{mpif_prompt22}
\end{equation}
Stacking the modality-specific prompts yields
$\mathbf{P}=\left[\mathbf{P}_{\text {rgb }}, \mathbf{P}_{\text {th }}, \mathbf{P}_{\text {polar }}\right]$.
These prompts serve as compact semantic carriers for BEV feature refinement. We perform geometry-guided prompt attention using the LiDAR BEV features as query tokens, which attend only to the modality prompts, thereby yielding a prompt-conditioned BEV feature map:

\begin{equation}            
\mathbf{F}_{\text {attn }}=\operatorname{Softmax}\left(\frac{\tilde{\mathbf{F}}_{l}\left(\mathbf{W}_{K} \mathbf{P}\right)^{\top}}{\sqrt{D / h}}\right) \mathbf{W}_{V} \mathbf{P},
   \label{mpif_prompt_attn}
\end{equation}
where $D$ denotes the embedding dimension and $h$ is the number of attention heads. 
To preserve geometric dominance while enabling adaptive semantic enhancement, we employ residual modulation:
\begin{equation}            
\begin{array}{c}
\mathbf{M}=\sigma\left(\gamma\left(\mathbf{F}_{a t t n}\right)\right), \\
\mathbf{F}_{f}=\tilde{\mathbf{F}}_{l}+\mathbf{M} \odot \tilde{\mathbf{F}}_{l},
\end{array}
   \label{mpif_res}
\end{equation}
where $\gamma(\cdot)$ is a $1\times1$ convolution and $\sigma(\cdot)$ is the sigmoid function.
This formulation allows prompts to adaptively reweight LiDAR BEV features instead of directly overriding them, ensuring that geometric structure remains the primary representation basis.

\begin{table}[t]
\centering
\setlength{\tabcolsep}{10pt}  % 列间距
\renewcommand{\arraystretch}{1.1}  % 行高
\caption{Results on different lighting conditions on PanoMMOcc.}
\vskip-2ex
\begin{tabular}{c|c|cc}
    \toprule
     \multirow{2}{*}{\#}& \multirow{2}{*}{Modality} &Day  & Night \\
    &&\multicolumn{2}{c}{mIoU$\uparrow$ } \\
    \midrule
    \multirow{4}{*}{VoxelHound} 
    %& C & 6.93 & 3.52 \\
    & C & 6.47 & 3.46 \\
    & C+T+P & 6.85 & 4.07 \\
    & C+L & 22.56 & 19.17 \\
    & C+L+T+P & 23.34 & 18.68 \\
    \bottomrule
\end{tabular}
\label{tab:day_night}
\vskip-4ex
\end{table}

\section{Experiments}
\subsection{Experiment Setup}
\subsubsection{Datasets}
We evaluate on panoramic multimodal semantic occupancy benchmarks for quadruped 
platforms: PanoMMOcc. 
We select $42$ sequences for annotation and experiments, among which $30$ sequences are used for training and $12$ sequences for testing, comprising $16.8k$ consecutive frames in total. 
Following~\cite{wang2023openoccupancy,tian2023occ3d}, we sample keyframes with a temporal stride of 5 for training and evaluation.
The occupancy annotations are defined over a 3D volume of size $64 \times 64 \times 16$. 
The spatial range covers $[-12.8, 12.8]$ meters in the x–y plane and $[-2.4, 4.0]$ meters along the vertical (z) axis. 
Each voxel has a resolution of $0.4m\times0.4m\times0.4m$ and is assigned one of $12$ semantic categories.

\begin{table}[t]
\centering
\setlength{\tabcolsep}{10pt}  % 列间距
\renewcommand{\arraystretch}{1.1}  % 行高
\caption{Ablation on different components in VoxelHound.}
\vskip-2ex
\begin{tabular}{ccc|cc}
    \toprule
    \# & VJC & MIPF& IoU$\uparrow$& mIoU$\uparrow$  \\
    \midrule

    1 & && 46.03&22.74  \\
    2 & \Checkmark &&45.96&\underline{22.97}  \\
    3 & & \Checkmark &\underline{46.06}& 22.88 \\
    Ours & \Checkmark & \Checkmark&\textbf{46.08}& \textbf{23.34}  \\
    \bottomrule
\end{tabular}
\label{tab:componts}
\vskip-4ex
\end{table}

\subsubsection{Evaluation Metric}
Following previous methods, we use the mean Intersection over Union (mIoU) as our evaluation metric:
\begin{equation}            
\mathrm{mIoU}=\frac{1}{C_{\mathrm{cls}}} 
\sum_{c=1}^{C_{\mathrm{cls}}} 
\frac{TP_{i}}{TP_{i}+FP_{i}+FN_{i}},
\label{Metric}
\end{equation}
where $TP_{i}$, $TP_{i}$, $FP_{i}$ denote the number of true positive, false positive and false negative predictions for class $i$, and $C_{\mathrm{cls}}$ is the total number of semantic classes.

\subsubsection{Loss}
During training, we use existing losses from prior works~\cite{wang2023openoccupancy}. 
They are cross-entropy loss $\mathcal{L}_{\mathrm{ce}}$, lovasz-softmax loss $\mathcal{L}_{\mathrm{ls}}$~\cite{berman2018lovasz}, affinity loss $\mathcal{L}_{\mathrm{scal}}^{\text {geo}}$ and $\mathcal{L}_{\mathrm{scal}}^{\text {sem}}$~\cite{cao2022monoscene}.
Therefore, the overall loss function $\mathcal{L}_{\mathrm{occ}}$ can be derived as:
\begin{equation}            
\mathcal{L}_{\mathrm{occ}}=\mathcal{L}_{\mathrm{ce}}+\mathcal{L}_{\mathrm{ls}}+\mathcal{L}_{\mathrm{scal}}^{\text {geo}}+\mathcal{L}_{\mathrm{scal}}^{\text {sem}}.
   \label{loss}
\end{equation}

\subsubsection{Implementation Details}
Unless noted, we follow the official configuration~\cite{shi2024effocc} with minimal dataset-specific modifications. 
All experiments are conducted on $4$ NVIDIA RTX 3090 GPUs.
We adopt the AdamW optimizer with an initial learning rate of $4 \times 10^{-4}$ and a weight decay of $0.01$. 
The model is trained for 48 epochs. For each modality, the corresponding image encoder uses ResNet-18 as its backbone.

\begin{figure*}[!t]
    \centering
    \includegraphics[width=0.94\linewidth]{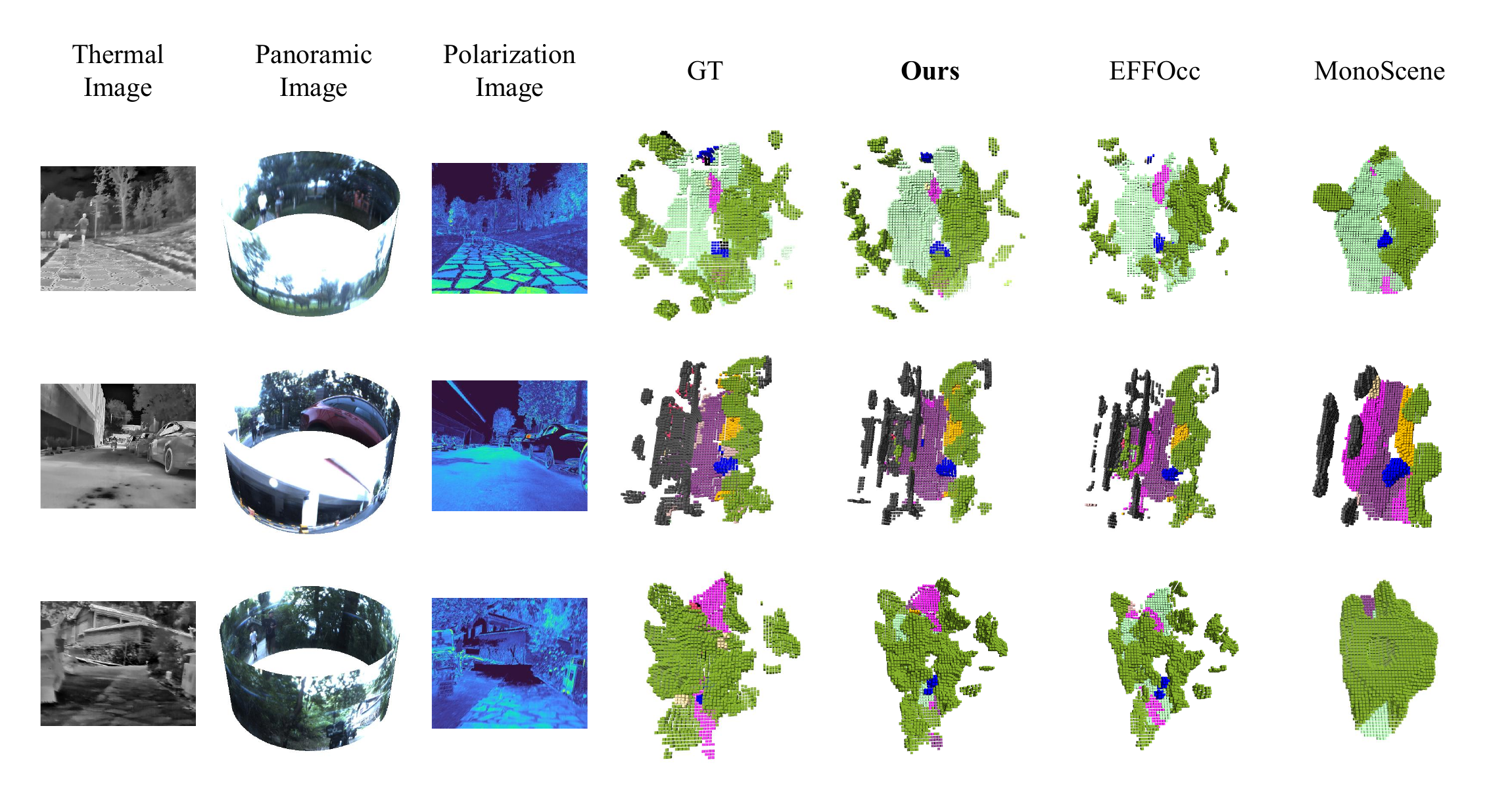}
    \vskip-2ex
    \caption{Qualitative comparison on PanoMMOcc. 
    From left to right: thermal image, panoramic image, polarization image, GT occupancy, VoxelHound (ours), EFFOcc~\cite{shi2024effocc}, and MonoScene~\cite{cao2022monoscene}.
    VoxelHound produces more complete scene structures and preserves finer semantic details in complex regions.}
    \label{fig:vis}
    \vskip-4ex
\end{figure*}

\begin{figure}[!t]
    \centering
    \includegraphics[width=\linewidth]{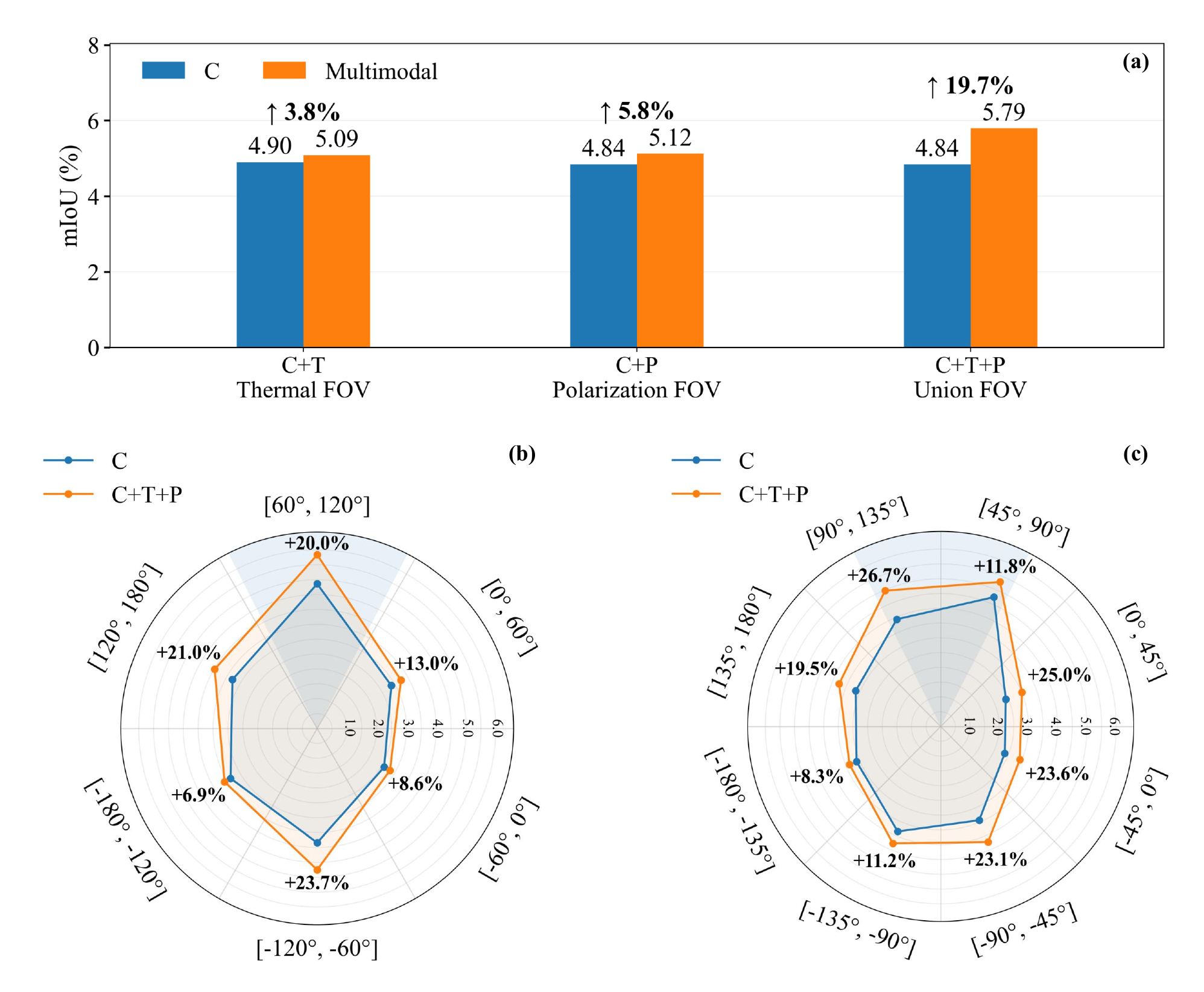} 
    \vskip-2ex
    \caption{FOV-aware evaluation of multimodal occupancy perception. 
    (a) Comparison between C and the corresponding multimodal inputs within the thermal, polarization, and union FOVs;
    (b-c) directional mIoU over all 3D voxel space, partitioned by platform-centered azimuth into six $60^\circ$ and eight $45^\circ$ sectors, respectively, with forward at the top.}
    \label{fig:fov_analysis}
    \vskip-4ex
\end{figure}

\subsection{Results and Analyses}
\subsubsection{Comparison to State-of-the-Art Methods}
To evaluate the effectiveness of the proposed VoxelHound, we conduct comprehensive experiments under different modality settings on PanoMMocc.
As shown in Tab.~\ref{tab:occ_performance},
we compare VoxelHound with several representative monocular and multimodal occupancy prediction methods.
Our full multimodal model VoxelHound (C+L+T+P) achieves the best overall performance, reaching $23.34\%$ mIoU and outperforming all competing methods.
Compared with MonoScene ($8.94\%$ mIoU), our method achieves a substantial improvement of $+14.40\%$ mIoU, demonstrating the strong advantage of multimodal sensing over camera-only perception.
Under the same C+L setting, VoxelHound achieves $22.87\%$ mIoU, outperforming EFFOcc-T ($19.18\%$ mIoU) by $+3.69\%$.
Furthermore, the full C+L+T+P model exceeds EFFOcc-T by $+4.16\%$ mIoU, indicating that thermal and polarization cues provide additional complementary information.
These results clearly demonstrate that the proposed architecture better exploits cross-modal complementary information.
We further conduct experiments under daytime and nighttime conditions to evaluate the robustness of different modality configurations under varying illumination, as shown in Tab.~\ref{tab:day_night}.
Furthermore, we present the qualitative experimental results of VoxelHound on the PanoMMOcc test set in Fig.~\ref{fig:vis}. 
The visualizations show that our method produces more complete and accurate occupancy predictions, especially in complex scenes and object boundaries.

\begin{table}[t]
\centering
%\small
\setlength{\tabcolsep}{7.5pt}  % 列间距
\renewcommand{\arraystretch}{1.1}  % 行高
\caption{Ablation study of prompt channel dimensions and attention heads in the MIPF module.}
\vskip-2ex
\begin{tabular}{cc|cc|c}
    \toprule
    $C_{pd}$ & $C_{nh}$ & Params. (M)$\downarrow$ & Mem. (MB)$\downarrow$  & mIoU$\uparrow$ \\
    \midrule
    \multirow{2}{*}{8} 
        & 4 & \multirow{2}{*}{43.50} & \multirow{2}{*}{386.97} & 23.13 \\
        & 8 &  &  & \textbf{23.34} \\
        \midrule
    \multirow{2}{*}{16} 
        & 4 & \multirow{2}{*}{43.51} & \multirow{2}{*}{387.02} & 22.91 \\
        & 8 &  &  & 22.94 \\
        \midrule
    \multirow{2}{*}{32} 
        & 4 & \multirow{2}{*}{43.54} & \multirow{2}{*}{387.11} & 22.67 \\
        & 8 &  &  & 22.48 \\
    \bottomrule
\end{tabular}
\label{tab:mipf}
\vskip-4ex
\end{table}

\subsubsection{Beyond-FOV Benefits of Multimodal Perception}
For fair FOV-specific evaluation, the pedestrian class is excluded because the data-collection operator consistently appears behind the platform and is therefore invisible to the forward-facing thermal and polarization cameras. 
As shown in Fig.~\ref{fig:fov_analysis}(a), C+T and C+P improve the matched-FOV mIoU by $3.8\%$ and $5.8\%$, respectively, while C+T+P yields a larger gain of $19.7\%$ over the union FOV.
Fig.~\ref{fig:fov_analysis}(b) and (c) further show consistent improvements across azimuth sectors, indicating that limited-FOV multimodal cues also benefit occupancy prediction beyond directly observed regions.

\subsection{Ablation Study}
We ablate VJC and MIPF in Tab.~\ref{tab:componts}. The baseline without either module achieves $22.74$ mIoU. Adding VJC or MIPF improves the performance to $22.97$ and $22.88$, respectively, while combining both further increases it to $23.34$ mIoU. 
These results verify the effectiveness of each module and demonstrate their complementary contributions when employed together.
Tabs.~\ref{tab:vjc} and~\ref{tab:mipf} further analyze their internal configurations. The best results are obtained with $C_{hd}=64$ for VJC and $C_{pd}=8$ with eight attention heads for MIPF, while introducing only marginal parameter and memory overhead.

\begin{table}[t]
\centering
\setlength{\tabcolsep}{10pt}  % 列间距
\renewcommand{\arraystretch}{1.1}  % 行高
\caption{Ablation study of different hidden channel dimensions for the VJC module.}
\vskip-2ex
\begin{tabular}{c|cc|c}
    \toprule
    $C_{hd}$ & Params. (M)$\downarrow$ & Mem. (MB)$\downarrow$  & mIoU$\uparrow$ \\
    \midrule
    1  & 43.46 & 386.83  & 23.12 \\
    4  & 43.46 & 386.84  & 23.01 \\
    8  & 43.47 & 386.84  & 22.90 \\
    16 & 43.47 & 386.86  & 23.11 \\
    32 & 43.48 & 386.89  & 22.88 \\
    64 & 43.50 & 386.97  & \textbf{23.34} \\
    128 & 43.56 & 387.21  & 22.68 \\
    \bottomrule
\end{tabular}
\label{tab:vjc}
\vskip-4ex
\end{table}

\section{Conclusion}
In this work, we introduce PanoMMOcc, 
a real-world panoramic multimodal occupancy dataset for quadruped robots, supporting fine-grained 3D scene understanding across diverse environments.
We further propose VoxelHound, a panoramic multimodal occupancy framework that efficiently aligns and fuses complementary cues while preserving geometric and semantic consistency. 
Together, PanoMMOcc and VoxelHound facilitate the systematic study of panoramic multimodal occupancy perception for quadruped robots.

In the future, we plan to investigate robust and efficient multimodal fusion under noisy or missing sensor inputs, incorporate temporal cues to improve occupancy consistency during quadruped locomotion, and extend VoxelHound to downstream embodied tasks, including 3D detection, mapping, and autonomous navigation.

\bibliographystyle{IEEEtran}
\bibliography{main}

@String(CVPR  = {IEEE Conf. Comput. Vis. Pattern Recog.})

@String(ICCV  = {Int. Conf. Comput. Vis.})

@String(ECCV  = {Eur. Conf. Comput. Vis.})

@String(NeurIPS = {Adv. Neural Inform. Process. Syst.})

@String(ACCV  = {Asian Conf. Comput. Vis.})

@String(CVPRW = {IEEE Conf. Comput. Vis. Pattern Recog. Worksh.})

@String(IJCAI = {IJCAI})

@String(ICIP  = {IEEE Int. Conf. Image Process.})

@String(ICASSP=	{ICASSP})

@String(CVPR  = {CVPR})

@String(ICCV  = {ICCV})

@String(ECCV  = {ECCV})

@String(NeurIPS = {NeurIPS})

@String(ACCV  = {ACCV})

@String(CVPRW = {CVPRW})

@String(ICIP  = {ICIP})

@article{ai2025survey,
  title={A Survey of Representation Learning, Optimization Strategies, and Applications for Omnidirectional Vision},
  author={Ai, Hao and Cao, Zidong and Wang, Lin},
  journal={International Journal of Computer Vision},
  year={2025}
}

@article{fernandez2018layouts,
  title={Layouts from panoramic images with geometry and deep learning},
  author={Clara Fernandez{-}Labrador and
                  Alejandro P{\'{e}}rez{-}Yus and
                  Gonzalo L{\'{o}}pez{-}Nicol{\'{a}}s and
                  Josechu J. Guerrero},
  journal={IEEE Robotics and Automation Letters},
  year={2018},
  publisher={IEEE}
}

@inproceedings{yang2019dula,
  title={{DuLa-Net:} {A} Dual-Projection Network for Estimating Room Layouts From a Single {RGB} Panorama},
  author={Yang, Shang-Ta and Wang, Fu-En and Peng, Chi-Han and Wonka, Peter and Sun, Min and Chu, Hung-Kuo},
  booktitle={CVPR},
  year={2019}
}

@inproceedings{shen2023disentangling,
  title={Disentangling orthogonal planes for indoor panoramic room layout estimation with cross-scale distortion awareness},
  author={Shen, Zhijie and others},
  booktitle={CVPR},
  year={2023}
}

@article{shen2024360,
  title={360 layout estimation via orthogonal planes disentanglement and multi-view geometric consistency perception},
  author={Shen, Zhijie and Lin, Chunyu and Zhang, Junsong and Nie, Lang and Liao, Kang and Zhao, Yao},
  journal={IEEE Transactions on Pattern Analysis and Machine Intelligence},
  year={2024},
  publisher={IEEE}
}

@inproceedings{guo2026event,
  title={Event-aided semantic scene completion},
  author={Guo, Shangwei and Shi, Hao and Wang, Song and Yin, Xiaoting and Yang, Kailun and Wang, Kaiwei},
  booktitle={ICASSP},
  year={2026}
}

@inproceedings{su2023gpr,
  title={{GPR-Net:} {Multi-view} Layout Estimation via a Geometry-aware Panorama Registration Network},
  author={Su, Jheng-Wei and Peng, Chi-Han and Wonka, Peter and Chu, Hung-Kuo},
  booktitle={CVPRW},
  year={2023}
}

@article{yang2019pass,
  title={{PASS:} {Panoramic} annular semantic segmentation},
  author={Kailun Yang and
                  Xinxin Hu and
                  Luis Miguel Bergasa and
                  Eduardo Romera and
                  Kaiwei Wang},
  journal={IEEE Transactions on Intelligent Transportation Systems},
  year={2020},
  publisher={IEEE}
}

@inproceedings{yang2020ds,
  title={{DS-PASS:} {Detail-sensitive} Panoramic Annular Semantic Segmentation through {SwaftNet} for Surrounding Sensing},
  author={Yang, Kailun and Hu, Xinxin and Chen, Hao and Xiang, Kaite and Wang, Kaiwei and Stiefelhagen, Rainer},
  booktitle={IV},
  year={2020}
}

@inproceedings{ma2021densepass,
  title={{DensePASS:} {Dense} Panoramic Semantic Segmentation via Unsupervised Domain Adaptation with Attention-Augmented Context Exchange},
  author={Ma, Chaoxiang and Zhang, Jiaming and Yang, Kailun and Roitberg, Alina and Stiefelhagen, Rainer},
  booktitle={ITSC},
  year={2021}
}

@inproceedings{zhang2022bending,
  title={Bending reality: Distortion-aware transformers for adapting to panoramic semantic segmentation},
  author={Zhang, Jiaming and Yang, Kailun and Ma, Chaoxiang and Rei{\ss}, Simon and Peng, Kunyu and Stiefelhagen, Rainer},
  booktitle={CVPR},
  year={2022}
}

@article{zhang2024behind,
  title={Behind every domain there is a shift: Adapting distortion-aware vision transformers for panoramic semantic segmentation},
  author={Jiaming Zhang and others},
  year={2024},
  journal={IEEE Transactions on Pattern Analysis and Machine Intelligence},
  publisher={IEEE}
}

@inproceedings{li2025voxdet,
  title={{VoxDet:} {Rethinking} {3D} Semantic Occupancy Prediction as Dense Object Detection},
  author={Li, Wuyang and Yu, Zhu and Alahi, Alexandre},
  booktitle={NeurIPS},
  year={2025}
}

@inproceedings{li2024sscbench,
  title={{SSCBench:} {A} Large-Scale {3D} Semantic Scene Completion Benchmark for Autonomous Driving},
  author={Yiming Li and others},
  booktitle={IROS},
  year={2024}
}

@article{wu2025synthetic_v2x,
  title={A Synthetic Benchmark for Collaborative {3D} Semantic Occupancy Prediction in {V2X} Autonomous Driving},
  author={Wu, Hanlin and others},
  journal={arXiv preprint arXiv:2506.17004},
  year={2025}
}

@article{min2025advancing_orad,
  title={Advancing Off-Road Autonomous Driving: The Large-Scale {ORAD-3D} Dataset and Comprehensive Benchmarks},
  author={Min, Chen and others},
  journal={arXiv preprint arXiv:2510.16500},
  year={2025}
}

@inproceedings{wu2025embodiedocc,
  title={{EmbodiedOcc:} {Embodied} {3D} Occupancy Prediction for Vision-based Online Scene Understanding},
  author={Wu, Yuqi and Zheng, Wenzhao and Zuo, Sicheng and Huang, Yuanhui and Zhou, Jie and Lu, Jiwen},
  booktitle={ICCV},
  year={2025}
}

@inproceedings{wang2025embodiedocc++,
  title={{EmbodiedOcc++:} {Boosting} Embodied {3D} Occupancy Prediction with Plane Regularization and Uncertainty Sampler},
  author={Wang, Hao and others},
  booktitle={MM},
  year={2025}
}

@inproceedings{behley2019semantickitti,
  title={{SemanticKITTI:} {A} Dataset for Semantic Scene Understanding of {LiDAR} Sequences},
  author={Behley, Jens and others},
  booktitle={ICCV},
  year={2019}
}

@inproceedings{wang2023openoccupancy,
  title={{OpenOccupancy:} {A} Large Scale Benchmark for Surrounding Semantic Occupancy Perception},
  author={Wang, Xiaofeng and others},
  booktitle={ICCV},
  year={2023}
}

@inproceedings{tian2023occ3d,
  title={{Occ3D:} {A} Large-Scale {3D} Occupancy Prediction Benchmark for Autonomous Driving},
  author={Tian, Xiaoyu and others},
  booktitle={NeurIPS},
  year={2023}
}

@article{lin2025one,
  title={One Flight Over the Gap: A Survey from Perspective to Panoramic Vision},
  author={Lin, Xin and others},
  journal={arXiv preprint arXiv:2509.04444},
  year={2025}
}

@article{ma2024licrocc,
  title={{LiCROcc:} {Teach} Radar for Accurate Semantic Occupancy Prediction Using {LiDAR} and Camera},
  author={Ma, Yukai and others},
  journal={IEEE Robotics and Automation Letters},
  year={2025},
  publisher={IEEE}
}

@inproceedings{ding2024radarocc,
  title={{RadarOcc:} {Robust} {3D} Occupancy Prediction with {4D} Imaging Radar},
  author={Ding, Fangqiang and Wen, Xiangyu and Zhu, Yunzhou and Li, Yiming and Lu, Chris Xiaoxuan},
  booktitle={NeurIPS},
  year={2024}
}

@inproceedings{pan2020semanticposs,
  title={{SemanticPOSS:} {A} Point Cloud Dataset with Large Quantity of Dynamic Instances},
  author={Pan, Yancheng and Gao, Biao and Mei, Jilin and Geng, Sibo and Li, Chengkun and Zhao, Huijing},
  booktitle={IV},
  year={2020}
}

@inproceedings{li2025occmamba,
  title={{OccMamba:} {Semantic} Occupancy Prediction with State Space Models},
  author={Li, Heng and Hou, Yuenan and Xing, Xiaohan and Ma, Yuexin and Sun, Xiao and Zhang, Yanyong},
  booktitle={CVPR},
  year={2025}
}

@inproceedings{li2023voxformer,
  title={{VoxFormer:} {Sparse} Voxel Transformer for Camera-Based {3D} Semantic Scene Completion},
  author={Yiming Li and others},
  booktitle={CVPR},
  year={2023}
}

@article{cui2025humanoid,
  title={Humanoid Occupancy: Enabling A Generalized Multimodal Occupancy Perception System on Humanoid Robots},
  author={Wei Cui and others},
  journal={arXiv preprint arXiv:2507.20217},
  year={2025}
}

@article{zhai2024wildocc,
  title={{WildOcc:} {A} benchmark for off-road {3D} semantic occupancy prediction},
  author={Zhai, Heng and Mei, Jilin and Min, Chen and Chen, Liang and Zhao, Fangzhou and Hu, Yu},
  journal={arXiv preprint arXiv:2410.15792},
  year={2024}
}

@article{zheng2026doracamom,
  title={Doracamom: {Joint} {3D} Detection and Occupancy Prediction with Multi-view {4D} Radars and Cameras for Omnidirectional Perception},
  author={Lianqing Zheng and others},
  journal={IEEE Transactions on Circuits and Systems for Video Technology},
  year={2026}
}

@article{zheng2024omnihd,
  title={{OmniHD-Scenes:} {A} next-generation multimodal dataset for autonomous driving},
  author={Lianqing Zheng and others},
  journal={IEEE Transactions on Pattern Analysis and Machine Intelligence},
  year={2026}
}

@article{wu2026unsocc,
  title={{UnsOcc:} {3D} Semantic Occupancy Prediction in Unstructured Scene via Rendering Fusion},
  author={Wu, Ye and Song, Ruiqi and Ding, Baiyong and Zeng, Nanxin and Cheng, Junjie and Ai, Yunfeng},
  journal={arXiv preprint arXiv:2606.03581},
  year={2026}
}

@inproceedings{wang2024occgen,
  title={{OccGen:} {Generative} Multi-modal {3D} Occupancy Prediction for Autonomous Driving},
  author={Wang, Guoqing and others},
  booktitle={ECCV},
  year={2024}
}

@inproceedings{wenke2025dur360bev,
  title={{Dur360BEV:} {A} Real-World 360-Degree Single Camera Dataset and Benchmark for Bird-Eye View Mapping in Autonomous Driving},
  author={Wenke E and others},
  booktitle={ICRA},
  year={2025}
}

@inproceedings{wang2025l2cocc,
  title={{L2COcc:} {Lightweight} Camera-Centric Semantic Scene Completion via Distillation of LiDAR Model},
  author={Wang, Ruoyu and others},
  booktitle={IROS},
  year={2025}
}

@article{liu2025articubevseg,
  title={{ArticuBEVSeg:} {Road} Semantic Understanding and Its Application in Bird's Eye View from Panoramic Vision System of Long Combination Vehicles},
  author={Liu, Weimin and Wang, Wenjun},
  journal={IEEE Robotics and Automation Letters},
  year={2025},
  publisher={IEEE}
}

@inproceedings{zhong2025omnisam,
  title={{OmniSAM:} {Omnidirectional} Segment Anything Model for {UDA} in Panoramic Semantic Segmentation},
  author={Zhong, Ding and others},
  booktitle={ICCV},
  year={2025}
}

@inproceedings{zheng2023look_neighbor,
  title={Look at the neighbor: Distortion-aware unsupervised domain adaptation for panoramic semantic segmentation},
  author={Zheng, Xu and Pan, Tianbo and Luo, Yunhao and Wang, Lin},
  booktitle={ICCV},
  year={2023}
}

@inproceedings{zheng2023both,
  title={Both style and distortion matter: Dual-path unsupervised domain adaptation for panoramic semantic segmentation},
  author={Zheng, Xu and Zhu, Jinjing and Liu, Yexin and Cao, Zidong and Fu, Chong and Wang, Lin},
  booktitle={CVPR},
  year={2023}
}

@inproceedings{zheng2024semantics,
  title={Semantics distortion and style matter: Towards source-free {UDA} for panoramic segmentation},
  author={Zheng, Xu and Zhou, Pengyuan and Vasilakos, Athanasios V. and Wang, Lin},
  booktitle={CVPR},
  year={2024}
}

@inproceedings{li2023sgat4pass,
  title={{SGAT4PASS:} {Spherical} Geometry-Aware Transformer for PAnoramic Semantic Segmentation},
  author={Li, Xuewei and Wu, Tao and Qi, Zhongang and Wang, Gaoang and Shan, Ying and Li, Xi},
  booktitle={IJCAI},
  year={2023}
}

@inproceedings{cao2024geometric,
  title={Geometric exploitation for indoor panoramic semantic segmentation},
  author={Cao Dinh, Duc and Kim, Seok Joon and Cho, Kyusung},
  booktitle={NeurIPS},
  year={2024}
}

@inproceedings{zhang2024goodsam,
  title={{GoodSAM:} {Bridging} Domain and Capacity Gaps via Segment Anything Model for Distortion-aware Panoramic Semantic Segmentation},
  author={Zhang, Weiming and Liu, Yexin and Zheng, Xu and Wang, Lin},
  booktitle={CVPR},
  year={2024}
}

@article{zhang2024goodsam++,
  title={{GoodSAM++:} {Bridging} Domain and Capacity Gaps via Segment Anything Model for Panoramic Semantic Segmentation},
  author={Zhang, Weiming and Liu, Yexin and Zheng, Xu and Wang, Lin},
  journal={arXiv preprint arXiv:2408.09115},
  year={2024}
}

@inproceedings{wei2024onebev,
  title={{OneBEV:} {Using} One Panoramic Image for Bird's-Eye-View Semantic Mapping},
  author={Wei, Jiale and Zheng, Junwei and Liu, Ruiping and Hu, Jie and Zhang, Jiaming and Stiefelhagen, Rainer},
  booktitle={ACCV},
  year={2024}
}

@inproceedings{teng2024360bev,
  title={{360BEV:} {Panoramic} semantic mapping for indoor bird's-eye view},
  author={Teng, Zhifeng and others},
  booktitle={WACV},
  year={2024}
}

@article{xu2025mamba4pass,
  title={{Mamba4PASS:} {Vision} Mamba for PAnoramic Semantic Segmentation},
  author={Xu, Jiayue and Xu, Chao and Zhao, Jianping and Han, Cheng and Li, Hua},
  journal={Displays},
  year={2025},
  publisher={Elsevier}
}

@article{zheng2024360sfuda++,
  title={{360SFUDA++:} {Towards} Source-Free {UDA} for Panoramic Segmentation by Learning Reliable Category Prototypes},
  author={Zheng, Xu and Zhou, Peng Yuan and Vasilakos, Athanasios V. and Wang, Lin},
  journal={IEEE Transactions on Pattern Analysis and Machine Intelligence},
  year={2024},
  publisher={IEEE}
}

@article{hu2022distortion,
  title={Distortion convolution module for semantic segmentation of panoramic images based on the image-forming principle},
  author={Hu, Xing and An, Yi and Shao, Cheng and Hu, Huosheng},
  journal={IEEE Transactions on Instrumentation and Measurement},
  year={2022},
  publisher={IEEE}
}

@article{orhan2022semantic,
  title={Semantic segmentation of outdoor panoramic images},
  author={Orhan, Semih and Bastanlar, Yalin},
  journal={Signal, Image and Video Processing},
  year={2022},
  publisher={Springer}
}

@inproceedings{tan2025dasc,
  title={{DASC-SPT:} {Towards} Self-Supervised Panoramic Semantic Segmentation},
  author={Tan, Tianlong and Chen, Bin and Cao, Hongliang and Yan, Chenggang and Ma, Yike and Dai, Feng},
  booktitle={WACV},
  year={2025}
}

@inproceedings{lan2025deformable,
  title={Deformable Spherical Geometry Transformer For Panoramic Semantic Segmentation},
  author={Lan, Boyang and Yang, Li and Xu, Mai and Jiang, Lai and Wang, Yufeng},
  booktitle={ICIP},
  year={2025}
}

@article{zhang2025humanoidpano,
  title={{HumanoidPano:} {Hybrid} Spherical {panoramic-LiDAR} Cross-Modal Perception for Humanoid Robots},
  author={Qiang Zhang and others},
  journal={arXiv preprint arXiv:2503.09010},
  year={2025}
}

@inproceedings{lv2026gau_occ,
  title={{Gau-Occ:} {Geometry-completed} Gaussians for Multi-Modal {3D} Occupancy Prediction},
  author={Lv, Chengxin and Li, Yihui and Yang, Hongyu and Wang, YunHong},
  booktitle={CVPR},
  year={2026}
}

@article{zhu2026visual_geometry_evidence,
  title={From Visual Geometry Evidence to Embodied Semantic Occupancy Memory},
  author={Hu Zhu and others},
  journal={arXiv preprint arXiv:2607.05543},
  year={2026}
}

@inproceedings{guttikonda2024single,
  title={Single frame semantic segmentation using multi-modal spherical images},
  author={Guttikonda, Suresh and Rambach, Jason},
  booktitle={WACV},
  year={2024}
}

@inproceedings{yogamani2024fisheyebevseg,
  title={{FisheyeBEVSeg:} {Surround} View Fisheye Cameras based Bird’s-Eye View Segmentation for Autonomous Driving},
  author={Yogamani, Senthil and Unger, David and Narayanan, Venkatraman and Kumar, Varun Ravi},
  booktitle={CVPRW},
  year={2024}
}

@inproceedings{ma2024cotr,
  title={{COTR:} {Compact} Occupancy TRansformer for Vision-Based {3D} Occupancy Prediction},
  author={Ma, Qihang and Tan, Xin and Qu, Yanyun and Ma, Lizhuang and Zhang, Zhizhong and Xie, Yuan},
  booktitle={CVPR},
  year={2024}
}

@inproceedings{samani2023f2bev,
  title={{F2BEV:} {Bird's} Eye View Generation from Surround-View Fisheye Camera Images for Automated Driving},
  author={Ekta U. Samani and
                  Feng Tao and
                  Dasari Harshavardhan Reddy and
                  Sihao Ding and
                  Ashis G. Banerjee},
  booktitle={IROS},
  year={2023}
}

@article{ming2025occfusion,
  title={{OccFusion:} {Multi-sensor} Fusion Framework for {3D} Semantic Occupancy Prediction},
  author={Ming, Zhenxing and Berrio, Julie Stephany and Shan, Mao and Worrall, Stewart},
  journal={IEEE Transactions on Intelligent Vehicles},
  year={2025},
  publisher={IEEE}
}

@article{pan2024co_occ,
  title={{Co-Occ:} {Coupling} Explicit Feature Fusion With Volume Rendering Regularization for Multi-Modal {3D} Semantic Occupancy Prediction},
  author={Pan, Jingyi and Wang, Zipeng and Wang, Lin},
  journal={IEEE Robotics and Automation Letters},
  year={2024},
  publisher={IEEE}
}

@inproceedings{duan2025sdgocc,
  title={{SDGOCC:} {Semantic} and Depth-Guided Bird's-Eye View Transformation for {3D} Multimodal Occupancy Prediction},
  author={Zaipeng Duan and others},
  booktitle={CVPR},
  year={2025}
}

@inproceedings{shi2025oneocc,
  title={{OneOcc:} {Semantic} Occupancy Prediction for Legged Robots with a Single Panoramic Camera},
  author={Hao Shi and others},
  booktitle={CVPR},
  year={2026}
}

@inproceedings{wei2023surroundocc,
  title={{SurroundOcc:} {Multi-camera} {3D} Occupancy Prediction for Autonomous Driving},
  author={Wei, Yi and Zhao, Linqing and Zheng, Wenzhao and Zhu, Zheng and Zhou, Jie and Lu, Jiwen},
  booktitle={ICCV},
  year={2023}
}

@inproceedings{oh2025_3d_prototype,
  title={{3D} Occupancy Prediction with Low-Resolution Queries via Prototype-aware View Transformation},
  author={Oh, Gyeongrok and others},
  booktitle={CVPR},
  year={2025}
}

@inproceedings{huang2024gaussianformer,
  title={{GaussianFormer:} {Scene} as Gaussians for Vision-Based {3D} Semantic Occupancy Prediction},
  author={Huang, Yuanhui and Zheng, Wenzhao and Zhang, Yunpeng and Zhou, Jie and Lu, Jiwen},
  booktitle={ECCV},
  year={2024}
}

@inproceedings{huang2023tpvformer,
  title={Tri-perspective view for vision-based {3D} semantic occupancy prediction},
  author={Huang, Yuanhui and Zheng, Wenzhao and Zhang, Yunpeng and Zhou, Jie and Lu, Jiwen},
  booktitle={CVPR},
  year={2023}
}

@inproceedings{tong2023scene_as_occupancy,
  title={Scene as Occupancy},
  author={Wenwen Tong and others},
  booktitle={ICCV},
  year={2023}
}

@article{gao2022review,
  title={Review on panoramic imaging and its applications in scene understanding},
  author={Gao, Shaohua and Yang, Kailun and Shi, Hao and Wang, Kaiwei and Bai, Jian},
  journal={IEEE Transactions on Instrumentation and Measurement},
  year={2022},
  publisher={IEEE}
}

@inproceedings{chang2026denoise,
  title={Denoise and Align: {Towards} Source-Free {UDA} for Robust Panoramic Semantic Segmentation},
  author={Chang, Yaowen and Cao, Zhen and Zheng, Xu and Mi, Xiaoxin and Dong, Zhen},
  booktitle={CVPR},
  year={2026}
}

@inproceedings{sun2026kd360,
  title={{KD360-VoxelBEV:} {LiDAR} and 360-degree Camera Cross Modality Knowledge Distillation for Bird's-Eye-View Segmentation},
  author={Sun, Yixin and others},
  booktitle={WACV},
  year={2026}
}

@article{ma2026monocular,
  title={Monocular {3D} Occupancy Perception for Robots on Sidewalks via Hybrid {2D-3D} Learning},
  author={Ma, Yukai and others},
  journal={arXiv preprint arXiv:2606.19122},
  year={2026}
}

@article{guo2026humanoid_omniocc,
  title={{Humanoid-OmniOcc:} {Stereo-based} Full-View Occupancy Dataset for Embodied {AI}},
  author={Guo, Xianda and others},
  journal={arXiv preprint arXiv:2606.22971},
  year={2026}
}

@article{meng2025_3d_scene_geometry,
  title={{3D} indoor scene geometry estimation from a single omnidirectional image: A comprehensive survey},
  author={Meng, Ming and Zhu, Yonggui and Zhao, Yufei and Li, Zhaoxin and Zhu, Zhe},
  journal={Computational Visual Media},
  year={2025}
}

@inproceedings{huang2024selfocc,
  title={{SelfOcc:} {Self-supervised} Vision-Based {3D} Occupancy Prediction},
  author={Huang, Yuanhui and Zheng, Wenzhao and Zhang, Borui and Zhou, Jie and Lu, Jiwen},
  booktitle={CVPR},
  year={2024}
}

@article{zhu2026panoramic_scene_analysis,
  title={Panoramic Scene Analysis: A Survey from Distortion-Aware Engineering to Sphere-Native Foundation Modeling},
  author={Qinfeng Zhu and Lei Fan},
  journal={arXiv preprint arXiv:2606.27745},
  year={2026}
}

@article{li2025fishbev,
  title={{FishBEV:} {Distortion-resilient} Bird's Eye View Segmentation with Surround-View Fisheye Cameras},
  author={Li, Hang and Sheng, Dianmo and Dong, Qiankun and Wang, Zichun and Xu, Zhiwei and Li, Tao},
  journal={arXiv preprint arXiv:2509.13681},
  year={2025}
}

@inproceedings{cao2022monoscene,
  title={{MonoScene:} {Monocular} {3D} Semantic Scene Completion},
  author={Cao, Anh-Quan and de Charette, Raoul},
  booktitle={CVPR},
  year={2022}
}

@article{yu2023flashocc,
  title={{FlashOcc:} {Fast} and Memory-Efficient Occupancy Prediction via Channel-to-Height Plugin},
  author={Yu, Zichen and others},
  journal={arXiv preprint arXiv:2311.12058},
  year={2023}
}

@article{yang2025daocc,
  title={{DAOcc:} {3D} Object Detection Assisted Multi-Sensor Fusion for {3D} Occupancy Prediction},
  author={Yang, Zhen and others},
  journal={IEEE Transactions on Circuits and Systems for Video Technology},
  year={2026},
  publisher={IEEE}
}

@inproceedings{scaramuzza2006toolbox,
  title={A toolbox for easily calibrating omnidirectional cameras},
  author={Scaramuzza, Davide and Martinelli, Agostino and Siegwart, Roland},
  booktitle={IROS},
  year={2006}
}

@inproceedings{luo2025omnidirectional,
  title={Omnidirectional Multi-Object Tracking},
  author={Luo, Kai and others},
  booktitle={CVPR},
  year={2025}
}

@inproceedings{yang2019can,
  title={Can we {PASS} beyond the field of view? {Panoramic} annular semantic segmentation for real-world surrounding perception},
  author={Yang, Kailun and others},
  booktitle={IV},
  year={2019}
}

@article{kim2024pair360,
  title={{PAIR360:} {A} Paired Dataset of High-Resolution 360{\textdegree} Panoramic Images and {LiDAR} Scans},
  author={Kim, Geunu and Kim, Daeho and Jang, Jaeyun and Hwang, Hyoseok},
  journal={IEEE Robotics and Automation Letters},
  year={2024},
  publisher={IEEE}
}

@inproceedings{wu2025quadreamer,
  title={{QuaDreamer:} {Controllable} Panoramic Video Generation for Quadruped Robots},
  author={Wu, Sheng and others},
  booktitle={CoRL},
  year={2025}
}

@inproceedings{chen2024360+,
  title={{360+x:} {A} Panoptic Multi-modal Scene Understanding Dataset},
  author={Chen, Hao and Hou, Yuqi and Qu, Chenyuan and Testini, Irene and Hong, Xiaohan and Jiao, Jianbo},
  booktitle={CVPR},
  year={2024}
}

@inproceedings{dong2024panocontext,
  title={{PanoContext-Former:} {Panoramic} total scene understanding with a transformer},
  author={Dong, Yuan and Fang, Chuan and Bo, Liefeng and Dong, Zilong and Tan, Ping},
  booktitle={CVPR},
  year={2024}
}

@inproceedings{shi2024effocc,
  title={{EFFOcc:} {Learning} Efficient Occupancy Networks from Minimal Labels for Autonomous Driving},
  author={Shi, Yining and others},
  booktitle={IROS},
  year={2025}
}

@inproceedings{berman2018lovasz,
  title={The lov{\'a}sz-softmax loss: A tractable surrogate for the optimization of the intersection-over-union measure in neural networks},
  author={Berman, Maxim and Triki, Amal Rannen and Blaschko, Matthew B},
  booktitle={CVPR},
  year={2018}
}

@inproceedings{ha2017mfnet,
  title={{MFNet:} {Towards} real-time semantic segmentation for autonomous vehicles with multi-spectral scenes},
  author={Ha, Qishen and Watanabe, Kohei and Karasawa, Takumi and Ushiku, Yoshitaka and Harada, Tatsuya},
  booktitle={IROS},
  year={2017}
}

@article{xiang2021polarization,
  title={Polarization-driven semantic segmentation via efficient attention-bridged fusion},
  author={Xiang, Kaite and Yang, Kailun and Wang, Kaiwei},
  journal={Optics Express},
  year={2021},
  publisher={Optical Society of America}
}

@article{ming2024occfusion,
  title={{OccFusion:} {Multi-sensor} Fusion Framework for {3D} Semantic Occupancy Prediction},
  author={Ming, Zhenxing and Berrio, Julie Stephany and Shan, Mao and Worrall, Stewart},
  journal={IEEE Transactions on Intelligent Vehicles},
  year={2025},
  publisher={IEEE}
}

@article{li2026scaling,
  title={Scaling up occupancy-centric driving scene generation: Dataset and method},
  author={Li, Bohan and others},
  journal={IEEE Transactions on Pattern Analysis and Machine Intelligence},
  year={2026},
  publisher={IEEE}
}

@inproceedings{yu2024monocular_occupancy,
  title={Monocular occupancy prediction for scalable indoor scenes},
  author={Yu, Hongxiao and Wang, Yuqi and Chen, Yuntao and Zhang, Zhaoxiang},
  booktitle={ECCV},
  year={2024}
}

@ARTICLE{li2026occdistill,
  author={Li, Ruikai and others},
  journal={IEEE Transactions on Multimedia}, 
  title={{OccDistill:} {Distilling} Multi-modal Knowledge into Camera-based 3D Occupancy Networks for Autonomous Driving}, 
  year={2026}
}

\newpage

\clearpage
\setcounter{page}{1}
\appendices
\counterwithin{figure}{section}
\counterwithin{table}{section}

\section{Dataset Construction}
\label{sec:dataset_construction}

\begin{figure*}[!t]
    \centering
    \includegraphics[width=0.95\linewidth]{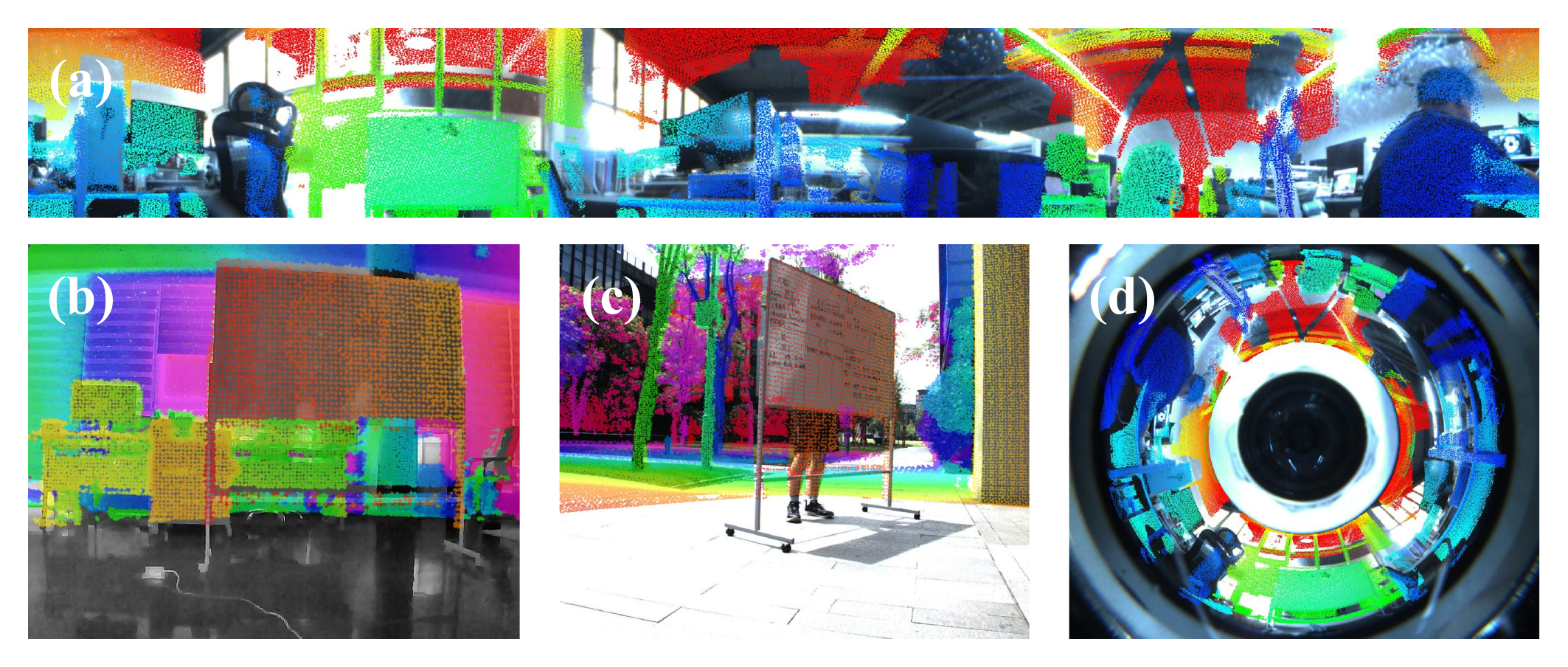} 

    \vskip -2ex
    \caption{Qualitative visualization of LiDAR--camera extrinsic calibration. 
LiDAR point clouds are projected onto different camera views using the calibrated extrinsic parameters: 
(a) equirectangularly unwrapped PAL image, 
(b) thermal image, 
(c) polarization image, and 
(d) original PAL image. 
The consistent alignment between projected points and image structures demonstrates the accuracy of the calibration.}
\vskip -2ex
    \label{fig:vis_calibration}
\end{figure*}

\subsection{Sensor Systems}
\label{sec:sensor_system}
Our panoramic multimodal acquisition platform is built on a Unitree Go2 quadruped robot and integrates a PAL panoramic camera, a MiD-360 LiDAR, a N-Driver P302B thermal camera, and a LUCID TRI050S-QC polarization camera for synchronized data collection in real-world environments. Here, we provide the details of polarization representation used in our dataset.
The polarization camera captures four intensity images at different polarization angles, \textit{i.e.}, $0^{\circ}$, $45^{\circ}$, $90^{\circ}$, and $135^{\circ}$.
The polarization state of light can be described by the Stokes vector:
\begin{equation}
\mathbf{S} = [S_0, S_1, S_2, S_3]^T ,
\label{supp_stoke_vectors}
\end{equation}
where $S_0$ denotes the total intensity, $S_1$ and $S_2$ describe the linear polarization components, and $S_3$ corresponds to the circular polarization component.
Since circular polarization is not considered in this work, only $S_0$, $S_1$, and $S_2$ are used:
\begin{equation}
S_0 = I_0 + I_{90}, \quad
S_1 = I_0 - I_{90}, \quad
S_2 = I_{45} - I_{135}.
\label{supp_stoke}
\end{equation}
where $I_0$, $I_{45}$, $I_{90}$, and $I_{135}$ are the intensity images captured at the corresponding polarization angles.
Based on these Stokes parameters, the Degree of Linear Polarization (DoLP) and Angle of Linear Polarization (AoLP) are computed as:
\begin{equation}
\text{DoLP} = \frac{\sqrt{S_1^2 + S_2^2}}{S_0}, \quad
\text{AoLP} = \frac{1}{2}\arctan\left(\frac{S_2}{S_1}\right).
\label{supp_dolp_aolp}
\end{equation}
DoLP measures the proportion of linearly polarized light, while AoLP represents the orientation of the polarization direction.

\subsection{Time Synchronization}
\label{time_syn}
To ensure temporal consistency across multimodal data streams, we adopt a host-based time synchronization scheme. 
A single host PC serves as the central acquisition unit and provides the global time reference for all sensors. 
The LiDAR and polarization camera are connected through a Gigabit Ethernet switch, while the panoramic and thermal cameras are directly connected to the PC.
All sensors are accessed and controlled through their corresponding device SDKs. Before each data collection session, the internal clocks of all devices are synchronized with the host PC's time reference. 
The LiDAR transmits point cloud packets with internal timestamps, which are recorded and aligned using the host system clock at the time of reception. This process ensures consistent temporal referencing of LiDAR measurements. For all cameras, frame-level timestamps are obtained directly from the device drivers and referenced to the synchronized host system time. 
As a result, all captured frames are associated with timestamps within the same time domain.

During post-processing, multimodal data streams are aligned according to these timestamps to construct synchronized observations for downstream perception tasks.

\definecolor{ncar}{RGB}{255, 158, 0}
\definecolor{nmotorcycle}{RGB}{255, 61, 99}
\definecolor{nbicycle}{RGB}{219, 11, 32}
\definecolor{npedestrain}{RGB}{0, 0, 230}
\definecolor{ntraffic_road}{RGB}{128 ,64 ,128}
\definecolor{nsidewalk}{RGB}{244, 35, 232}
\definecolor{nterrain}{RGB}{152, 251, 152}
\definecolor{nvegetation}{RGB}{107 ,142 ,35}
\definecolor{nbuilding}{RGB}{70, 70, 70}
\definecolor{nbarrier}{RGB}{190 ,153 ,153}
\definecolor{nmanmade}{RGB}{222, 184, 135}
\definecolor{npillar}{RGB}{153, 153, 153}

\captionsetup{font=small}  % 设置标题字体大小为小号

\setlength{\tabcolsep}{1.4pt}  % 列间距

\begin{table*}[t]
\centering
\renewcommand\arraystretch{1.1}
\caption{Performance of the proposed VoxelHound model across different scene categories on the established PanoMMOcc dataset.
}
\vskip-2ex
\begin{adjustbox}{width=0.95\textwidth,center}
% \resizebox{0.9\textwidth}{!}{
\begin{tabular}{c|c|c|cccccccccccc}
\hline
\noalign{\smallskip}
Scenes & Modality & mIoU & \rotatebox{90}{\textcolor{ncar}{$\blacksquare$} car}     & \rotatebox{90}{\textcolor{nmotorcycle}{$\blacksquare$} motorcycle}    & \rotatebox{90}{\textcolor{nbicycle}{$\blacksquare$} bicycle} & \rotatebox{90}{\textcolor{npedestrain}{$\blacksquare$} pedestrian} & \rotatebox{90}{\textcolor{ntraffic_road}{$\blacksquare$} driveable surface} & \rotatebox{90}{\textcolor{nsidewalk}{$\blacksquare$} sidewalk} & \rotatebox{90}{\textcolor{nterrain}{$\blacksquare$} terrain} & \rotatebox{90}{\textcolor{nvegetation}{$\blacksquare$} vegetation} & \rotatebox{90}{\textcolor{nbuilding}{$\blacksquare$} building} & \rotatebox{90}{\textcolor{nbarrier}{$\blacksquare$} barrier} & \rotatebox{90}{\textcolor{nmanmade}{$\blacksquare$} mannade} & \rotatebox{90}{\textcolor{npillar}{$\blacksquare$} pillar}    \\
\noalign{\smallskip}
\hline
\noalign{\smallskip}

\multirow{2}{*}{Urban}&C&4.59&0.00&13.20&0.00&16.36&14.97&2.09&4.00&4.25&0.00&0.13&0.10&0.00 \\
&C+L+T+P&16.96&18.76&27.66&0.00&41.71&26.05&7.99&11.27&29.89&0.00&21.67&7.94&10.56\\ \hline

\multirow{2}{*}{Residential}&C&4.14&2.37&0.17&0.00&19.68&21.96&0.30&0.29&2.17&0.58&1.83&0.28&0.00 \\
&C+L+T+P&20.30&31.47&10.40&0.00&49.93&37.67&3.45&11.63&35.17&42.61&5.25&2.53&13.49\\ \hline

\multirow{2}{*}{Campus}&C&4.97&0.19&1.07&0.00&25.61&25.05&1.49&2.45&0.55&0.90&1.82&0.00&0.53 \\
&C+L+T+P&26.50&29.13&22.35&0.00&55.08&43.35&7.43&20.13&33.47&49.11&27.31&0.36&30.23\\ \hline

\multirow{2}{*}{Green Spaces}&C&3.60&0.00&0.00&0.00&22.80&0.00&1.87&11.35&5.95&0.37&0.84&0.00&0.01 \\
&C+L+T+P&15.35&0.00&0.00&0.00&51.48&0.00&15.78&36.58&36.87&15.27&7.73&11.85&8.60\\ \hline

\multirow{2}{*}{Forest}&C&3.86&0.00&0.00&0.00&24.50&0.00&4.70&2.25&14.90&0.00&0.00&0.00&0.00 \\
&C+L+T+P&13.10&0.00&3.16&0.00&54.11&0.00&10.68&11.13&37.32&31.64&0.00&0.34&8.84\\ \hline

\multirow{2}{*}{Rural}&C&3.39&0.00&0.00&0.00&17.95&16.06&0.00&3.47&2.86&0.16&0.22&0.00&0.00 \\
&C+L+T+P&15.48&1.39&0.00&0.00&47.57&35.20&2.04&15.06&26.67&24.61&12.42&6.48&14.30\\

\noalign{\smallskip}
\hline
\end{tabular}
\label{tab:occ_scene_performance}
\end{adjustbox}
\vskip-4ex
\end{table*}
\setlength{\tabcolsep}{1.4pt}

\subsection{Calibration}
\label{calib}
To ensure geometric consistency across heterogeneous modalities, we calibrate all sensors in a unified LiDAR-centric coordinate system. 
The calibration pipeline consists of camera intrinsic calibration and LiDAR--camera extrinsic calibration.

\noindent \textbf{Camera Intrinsic Calibration.}
We independently calibrate the panoramic, thermal, and polarization cameras using checkerboard patterns captured from diverse viewpoints. 
For thermal and polarization cameras, we adopt the standard pinhole model with radial and tangential distortion. 
For the Panoramic Annular Lens (PAL) camera, we use the generic Taylor (OCam) model~\cite{scaramuzza2006toolbox} to handle its strong nonlinear omnidirectional projection. 
All intrinsic parameters are estimated by minimizing the reprojection error of checkerboard corners and are fixed for the entire dataset.

\noindent \textbf{LiDAR--Camera Extrinsic Calibration.}
After intrinsic calibration, we estimate the rigid transformation between the LiDAR frame $\mathcal{F}_{L}$ and each camera frame $\mathcal{F}_{C}$. 
Given a LiDAR point $\mathbf{p}_{L}$, its coordinate in the camera frame is computed as
\begin{equation}
    \mathbf{p}_{C}=\mathbf{R}_{L}^{C}\mathbf{p}_{L}+\mathbf{t}_{L}^{C},
    \label{supp_lidar_cam_transform}
\end{equation}
where $\mathbf{R}_{L}^{C}$ and $\mathbf{t}_{L}^{C}$ denote the LiDAR-to-camera rotation and translation, respectively.

We use a planar whiteboard with four well-defined corners for extrinsic calibration. 
The board is placed at multiple poses within the overlapping field of view of the LiDAR and cameras. 
For each frame, the four board corners are manually annotated in the image and selected from the LiDAR point cloud to avoid unreliable automatic detection under challenging modalities such as thermal and polarization imaging. 
Let $\mathbf{u}_{ij}$ denote the $j$-th annotated image corner in the $i$-th frame, and $\mathbf{P}^{L}_{ij}$ be its corresponding 3D LiDAR corner. 
The extrinsic parameters are optimized by minimizing the reprojection error:
\begin{equation}
    \min_{\mathbf{R}_{L}^{C}, \mathbf{t}_{L}^{C}}
    \sum_i \sum_{j=1}^{4}
    \left\|
    \mathbf{u}_{ij} -
    \pi_C\left(\mathbf{R}_{L}^{C}\mathbf{P}^{L}_{ij}+\mathbf{t}_{L}^{C}\right)
    \right\|^2 ,
    \label{supp_extrinsic_optimization}
\end{equation}
where $\pi_C(\cdot)$ denotes the calibrated projection function of the corresponding camera, \textit{i.e.}, the pinhole model for thermal and polarization cameras and the Taylor/OCam model for the PAL camera. 
This procedure is independently applied to each camera, producing LiDAR-to-camera transformations $\{\mathbf{T}_{L}^{C_i}\}$ for all modalities.

\noindent \textbf{Calibration Visualization.}
We qualitatively verify the estimated extrinsic parameters by projecting LiDAR point clouds onto images from different camera modalities.
As shown in Fig.~\ref{fig:vis_calibration}(a), Fig.~\ref{fig:vis_calibration}(b), Fig.~\ref{fig:vis_calibration}(c), and Fig.~\ref{fig:vis_calibration}(d), the projected point clouds are well aligned with image structures across thermal, polarization, PAL, and unwrapped PAL views.

\subsection{Data Collection}
\label{data_coll}
PanoMMOcc is collected by a quadruped robot in diverse real-world outdoor environments under both daytime and nighttime conditions. 
The dataset covers \textbf{urban street scenes}, \textbf{residential areas}, \textbf{recreational green spaces}, \textbf{campus environments}, \textbf{rural areas}, and \textbf{forested regions}, spanning highly structured built environments to weakly structured natural terrains. 
These scenes exhibit diverse geometric layouts, semantic compositions, terrain conditions, and dynamic elements. 
Benefiting from the mobility and terrain adaptability of the quadruped robot, PanoMMOcc captures multimodal observations in complex environments that are difficult to access with wheeled platforms, supporting research on robust multimodal perception and embodied intelligence.

\subsection{Privacy Protection}
\label{privacy}
To ensure responsible data release, identifiable personal information is anonymized before publication. 
Specifically, human faces and vehicle license plates are detected and obscured with mosaic blurring, preventing sensitive information disclosure while retaining the visual data's utility for research.

\begin{figure*}[t]
    \centering
    \includegraphics[width=0.95\linewidth]{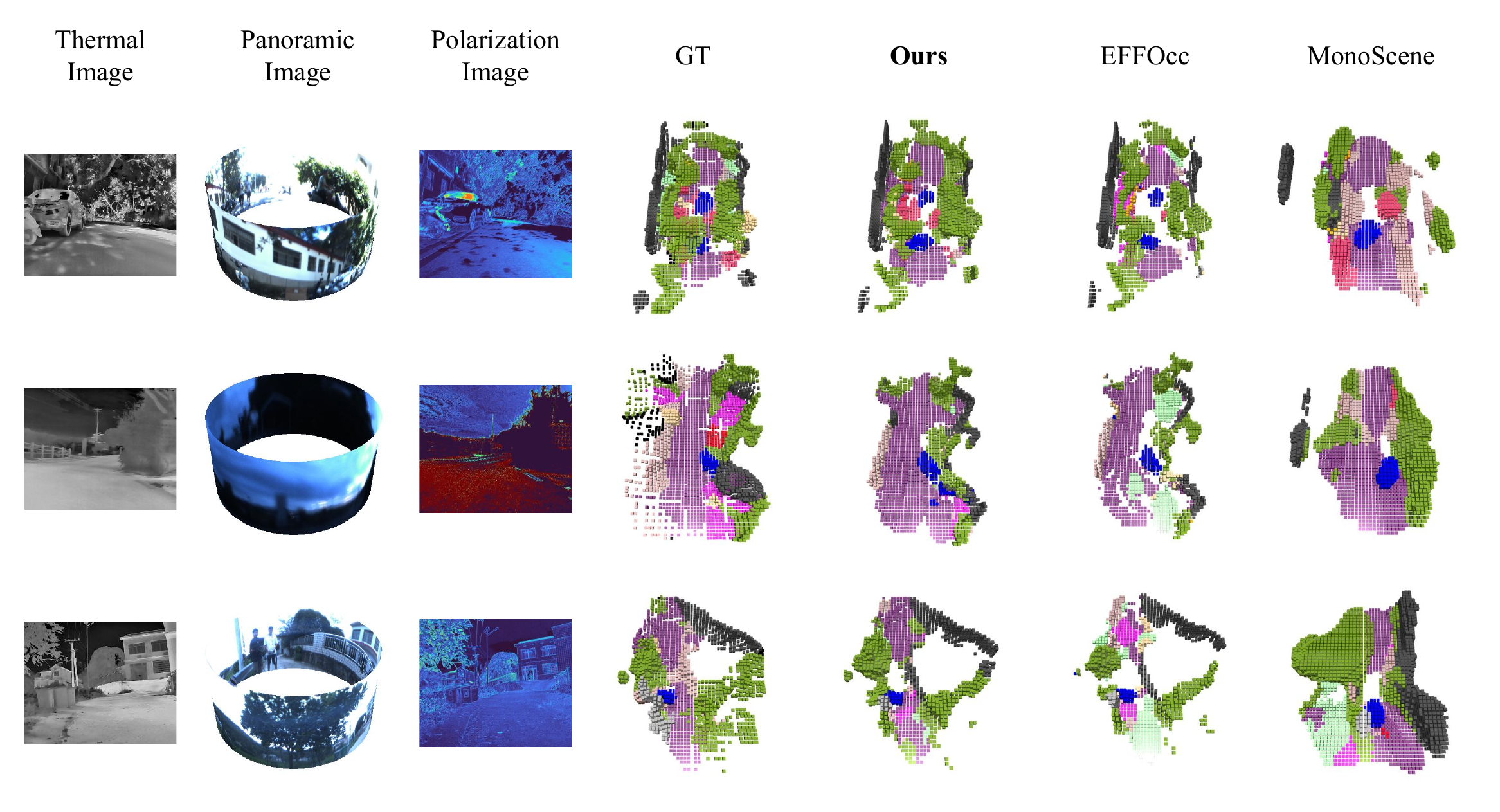}
    \vskip -2ex
    \caption{Qualitative comparison on PanoMMOcc under different environmental conditions. From left to right: thermal image, panoramic image, polarization image, GT occupancy, VoxelHound (ours), EFFOcc~\cite{shi2024effocc}, and MonoScene~\cite{cao2022monoscene}. From top to bottom: overexposed daytime, nighttime, and rural scenes. VoxelHound yields more coherent occupancy structures and more consistent semantic predictions under challenging conditions.}
    \vskip -4ex
    \label{fig:supp_vis}
\end{figure*}

\section{More Results}
\label{more_results}

\subsection{Performance Analysis under Different Scenes}
\label{per_ana_dif_sce}

Tab.~\ref{tab:occ_scene_performance} reports the scene-wise performance on PanoMMOcc. 
The full multimodal setting (C+L+T+P) consistently outperforms the camera-only setting (C), showing the benefit of combining panoramic images with LiDAR, thermal, and polarization cues. 
VoxelHound performs better in relatively structured, particularly residential and campus scenes, while performance drops in natural scenes such as green spaces, forests, and rural areas due to irregular terrain, dense vegetation, and ambiguous boundaries. 
These results highlight the importance of covering diverse scene types in PanoMMOcc and demonstrate the robustness of multimodal occupancy perception in complex outdoor environments.

\subsection{Voxel Resolution Analysis}
We further study the effect of voxel resolution while keeping the perception range, evaluation protocol, model architecture, and training hyperparameters unchanged. Specifically, we refine the voxel size from $0.4m$ to $0.2m$, corresponding to changing the voxel grid from ($64 \times 64 \times 16$) to ($128 \times 128 \times 32$). 
We report mIoU, Params, GFLOPs, FPS, and peak GPU memory measured on a single RTX 3090 with batch size $1$.

As shown in Tab.~\ref{tab:high_res}, the finer $128 \times 128 \times 32$ setting increases the computational cost from $84.47$ to $130.62$ GFLOPs, reduces the inference speed from $11.03$ to $10.57$ FPS, and increases the peak memory from $386.97$ to $393.99$ MB, while the parameter count changes only slightly from $43.50M$ to $43.61M$. Although finer voxelization provides more detailed spatial discretization, it also enlarges the prediction space and increases optimization difficulty. Consequently, the $0.2$ m setting achieves an mIoU of $18.28$, compared with $23.34$ under the $0.4$ m setting.
These results indicate that higher voxel resolution does not necessarily improve occupancy prediction under a fixed perception range. Considering the better accuracy--efficiency trade-off, we adopt $64 \times 64 \times 16$ as the default setting, while $128 \times 128 \times 32$ is more suitable for applications that prioritize finer geometric details.

\renewcommand{\arraystretch}{1.1}
\setlength{\tabcolsep}{1.5pt}
\begin{table}[t]
\centering
\caption{Effect of voxel resolution under a fixed perception range.}
\vskip-2ex
\begin{tabular}{cc|cccc|c}
    \toprule
    Resolution & Size & Params. (M)$\downarrow$ &GFLOPs$\downarrow$& Mem. (MB)$\downarrow$& FPS$\uparrow$ & mIoU$\uparrow$ \\
    \midrule
    $64 \times 64 \times 16$ & 0.4m & 43.50&84.47 & 386.97 & 11.03 & 23.34 \\
    $128 \times 128 \times 32$& 0.2m & 43.61&130.62 & 393.99 & 10.57 &18.28  \\

    \bottomrule
\end{tabular}
\label{tab:high_res}
\vskip-4ex
\end{table}

\subsection{More Visualizations}
\label{more_vis}
To further demonstrate the effectiveness of the proposed VoxelHound, we provide additional qualitative results under diverse environmental conditions. As shown in Fig.~\ref{fig:supp_vis}, we compare VoxelHound with representative occupancy prediction methods on challenging scenes, including overexposed daytime, nighttime, and rural environments. These scenes introduce significant perception difficulties due to strong illumination variations, low-light conditions, and irregular scene structures. Compared with other methods, VoxelHound produces more coherent occupancy structures and more consistent semantic predictions, particularly in regions affected by illumination degradation or sparse scene geometry. 
This improvement also benefits from the complementary multimodal sensing, where thermal and polarization cues provide additional information beyond RGB visual inputs, leading to more robust scene understanding.

\begin{figure*}[t]
    \centering
    \includegraphics[width=0.85\linewidth]{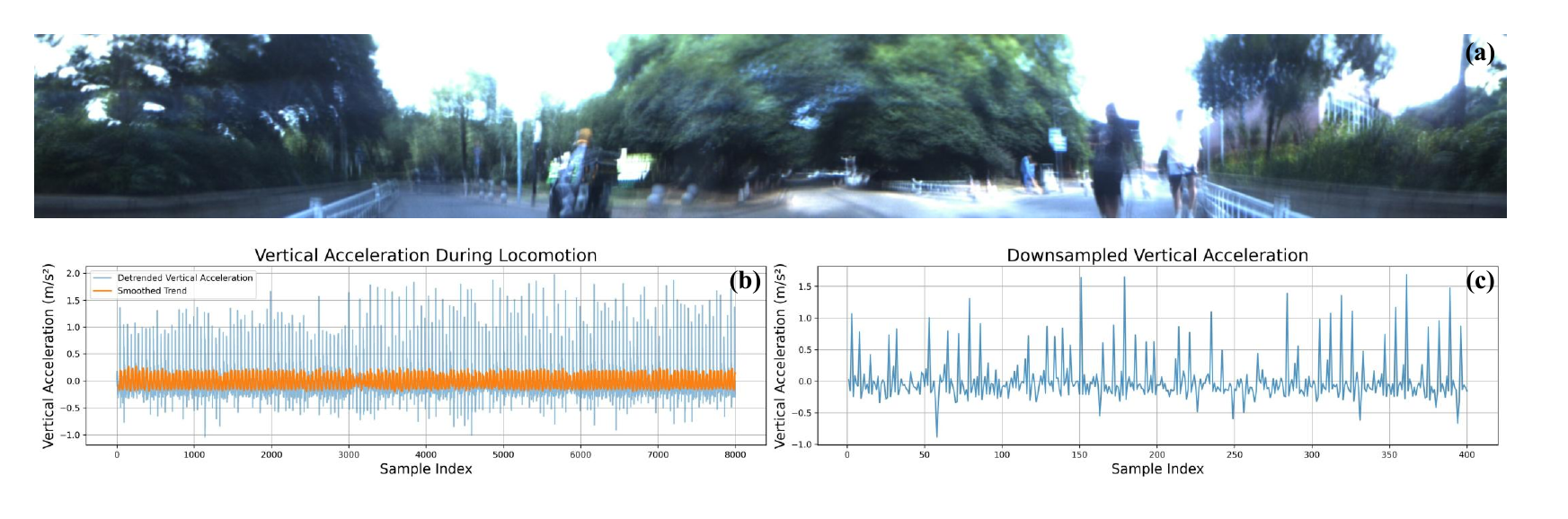}
    \vskip -2ex
    \caption{Example of gait-induced jitter in PanoMMOcc. 
    (a) Panoramic image with motion blur caused by gait-induced jitter during data acquisition.  
    (b) Vertical acceleration measured by the IMU integrated in the LiDAR during locomotion.
    (c) Downsampled vertical acceleration revealing the jitter pattern.}
    \vskip -4ex
    \label{fig:jitter}
\end{figure*}
\subsection{Gait-Induced Jitter Analysis}
\label{gait}
During data acquisition, the quadruped robot exhibits periodic body vibrations due to the contact of its legs with the ground, which differs from the relatively smooth motion of wheeled platforms. 
To analyze this gait-induced jitter, we use the vertical acceleration measured by the IMU integrated in the LiDAR. 
In particular, we focus on the vertical acceleration signal. Since the raw acceleration contains both motion-induced acceleration and the gravity component, we first remove the mean value to obtain the detrended vertical acceleration:
\begin{equation}
	a_{z}^{\prime}=a_{z}-\overline{a_{z}},
	\label{vjtter}
\end{equation}
where $a_{z}$ is the measured vertical acceleration and $\overline{a_{z}}$ denotes its mean value. 
The detrended signal $a_{z}^{\prime}$ mainly reflects locomotion-induced vibration.
Fig.\ref{fig:jitter}(a) shows an example panoramic image captured during robot motion, where noticeable motion blur can be observed due to body vibrations. Fig.\ref{fig:jitter}(b) plots the detrended vertical acceleration together with a smoothed curve obtained using a moving-average filter. The smoothed signal suppresses high-frequency components and highlights the low-frequency motion trend of the robot body, while the detrended acceleration reveals high-frequency oscillations caused by locomotion-induced impacts. 
Fig.\ref{fig:jitter}(c) further shows the downsampled detrended acceleration for clearer visualization of the vibration pattern.

From these observations, the detrended acceleration exhibits significant high-frequency oscillations, indicating pronounced vertical jitter during locomotion. Such oscillations originate from the periodic impacts between the robot's legs and the ground during the quadruped gait cycle. The resulting body vibration introduces noticeable motion blur in the captured images, which may degrade the quality of visual observations and pose challenges for downstream perception tasks.

\section{Discussions}
\label{Discussions}

\subsection{Limitations}
\label{limit}
Despite the advantages of the proposed dataset and perception framework, several limitations remain. 
Although PanoMMOcc covers a diverse set of environments, the overall dataset scale is still relatively limited compared with large-scale vehicle-based perception datasets. 
Expanding the dataset with more sequences, environments, and weather conditions would further improve its representativeness for real-world robotic applications.
In addition, indoor scenes are not extensively covered, which may limit the applicability of the dataset for certain robotic perception tasks.
%Future work will focus on expanding the dataset scale and incorporating more diverse environments, including indoor scenarios and challenging weather conditions.
Future work will investigate robust and efficient multimodal fusion under noisy or missing sensor inputs and incorporate temporal cues to improve occupancy consistency during quadruped locomotion

\begin{figure}[ht]
    \centering
    \includegraphics[width=0.98\linewidth]{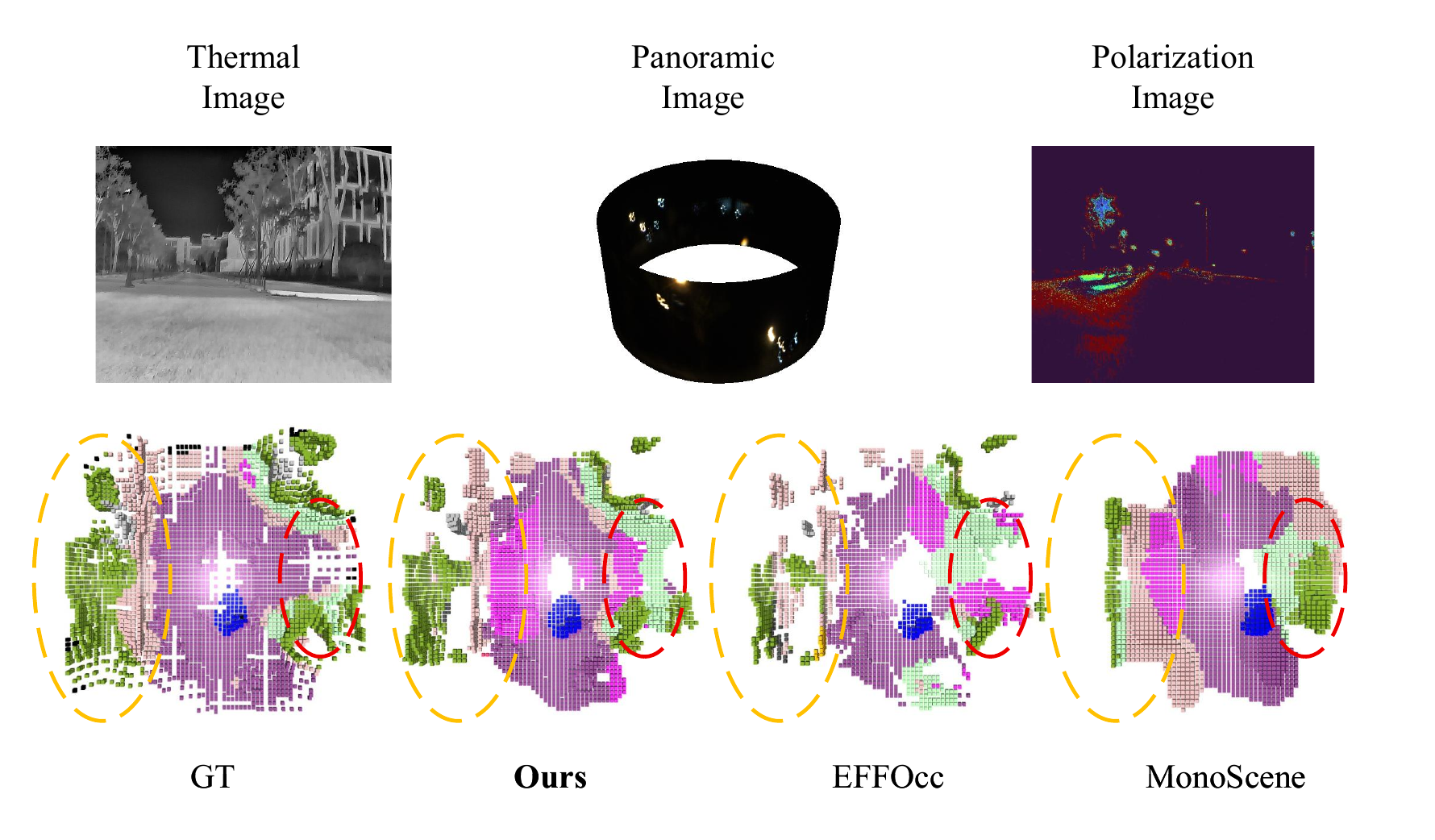}
    \vskip -2ex
    \caption{Failure case in an extremely dark nighttime scene. Due to the sparsity of long-range LiDAR observations, the predictions near the boundaries become sparse.}
    \vskip -4ex
    \label{fig:supp_fail_case}
\end{figure}
\subsection{Failure Case Analysis}
\label{fail_case}
Fig.\ref{fig:supp_fail_case} presents a representative failure case in an extremely dark nighttime environment. 
%In this scene, the predicted occupancy near the boundaries becomes relatively sparse, particularly for distant regions. This phenomenon is mainly caused by the reduced LiDAR point density at long ranges, as the Livox MID-360 LiDAR produces increasingly sparse observations with distance.
The predicted occupancy is sparse near scene boundaries, especially in distant regions. 
This phenomenon is mainly caused by the reduced LiDAR point density at long ranges, as the Livox MID-360 LiDAR produces increasingly sparse observations as distance increases.

Despite this limitation, our multimodal framework is still able to preserve the overall geometric structure of the scene. The predicted results of our method only exhibit sparsity in the far regions near the two sides. In contrast, EFFOcc (C+L) not only suffers from sparse predictions but also produces incorrect semantic categories in several regions due to the lack of complementary sensing cues. Meanwhile, MonoScene, which relies solely on visual inputs, fails to recover meaningful geometric structures under extremely low-light conditions.

Overall, this failure case highlights the challenges of occupancy perception in extremely dark environments and distant regions. At the same time, it also demonstrates the advantage of incorporating complementary sensing modalities, such as thermal and polarization information, which provide additional cues when RGB visual observations become unreliable.

%\subsection{Societal Impacts}

\end{document}